\documentclass{article}

\usepackage{PRIMEarxiv}

\usepackage{my}

\title{Semi-supervised 3D shape segmentation with multilevel consistency and part substitution}

\author{
  Chun-Yu Sun\\
  Tsinghua University \\
   \And
  Yu-Qi Yang\\
  Tsinghua University \\
   \And
  Hao-Xiang Guo\\
  Tsinghua University \\
   \And
  Peng-Shuai Wang \\
  Microsoft Research Asia \\
   \And
  Xin Tong\\
  Microsoft Research Asia \\
   \And
  Yang Liu \\
  Microsoft Research Asia \\
   \And
  Heung-Yeung Shum \\
  Tsinghua University \\
}

\begin{document}
\maketitle

\begin{abstract}
The lack of fine-grained 3D shape segmentation data is the main obstacle to developing learning-based 3D segmentation techniques. We propose an effective semi-supervised method for learning 3D segmentations from a few labeled 3D shapes and a large amount of unlabeled 3D data. For the unlabeled data, we present a novel \emph{multilevel consistency} loss to enforce consistency of  network predictions between  perturbed copies of a 3D shape at multiple levels: point-level, part-level, and hierarchical level. For the labeled data, we develop a simple yet effective part substitution scheme to augment the labeled 3D shapes with more structural variations to enhance training. Our method has been extensively validated on the task of 3D object semantic segmentation on PartNet and ShapeNetPart, and indoor scene semantic segmentation on ScanNet. It exhibits superior performance to  existing semi-supervised and unsupervised pre-training 3D approaches. Our code and trained models are publicly available at  \url{https://github.com/isunchy/semi_supervised_3d_segmentation}.
\end{abstract}

% keywords can be removed

\section{Introduction}\label{sec:intro}
Recognizing semantic parts of man-made 3D shapes is an essential task in computer vision and graphics. Man-made shapes often consist of fine-grained and semantic parts, many of which are small and hard to distinguish. Moreover, for 3D shapes within a shape category, the existence, geometry, and layout of semantic parts can often have large variations. As a result, obtaining accurate and consistent fine-grained segmentation for a shape category is challenging, even for human workers.

Recently, supervised learning approaches have been widely used in shape segmentation; they need sufficient labeled data. However, as there are not many large well-annotated 3D datasets, and the 3D data labeling process is costly and tedious, it is difficult to apply these methods to shape categories with limited labeled data. In this paper, we propose a novel semi-supervised approach for fine-grained 3D shape segmentation. Our method learns a deep neural network from a small set of segmented 3D point clouds and a large number of unlabeled 3D point clouds within a shape category, thus greatly reducing the workload of 3D data labeling.

We propose two novel schemes to efficiently utilize both unlabeled and labeled data for network training. For unlabeled data, inspired by the pixel-level consistency scheme used in semi-supervised image segmentation ~\cite{Ouali2020,Ke2020}, we propose a set of \emph{multilevel consistency} losses for measuring the consistency of network predictions between two perturbed copies of a 3D point cloud at the \emph{point-level}, \emph{part-level}, and \emph{hierarchical level}. Via the multilevel consistency, the data priors hidden in the unlabeled data can be learned by the network to good effect.  For the available labeled shapes, we present a simple yet effective multilevel \emph{part-substitution} algorithm to enrich the labeled data set by replacing parts with  semantically similar parts of other labeled data.  The algorithm is specially designed for 3D structured shapes, like chairs and tables, and it enhances the geometry and structural variation of the labeled data in a simple way, leading to a boost in network performance.

We evaluate the efficacy of our method on the task of 3D shape segmentation including object segmentation and indoor scene segmentation, by training the networks with different amounts of labeled data and unlabeled data. An ablation study further validates the significance of each type of consistency loss. Extensive experiments demonstrate the superiority of our method over  state-of-the-art semi-supervised and unsupervised 3D pretraining approaches.

\section{Related Work} \label{sec:related}

In this section, we briefly review  related 3D shape segmentation approaches and  shape synthesis techniques.

\subsection{Unsupervised 3D segmentation}
Early attempts at unsupervised segmentation focused on decomposing a single shape into meaningful geometric parts using clustering, graph cuts, or primitive fitting (see surveys in~\cite{Shamir2008,Rodrigues2018}). To obtain consistent segmentation within a shape category, a series of unsupervised co-segmentation works (see  surveys in~\cite{Xu2015a,Rodrigues2018}) proposed  exploiting geometrically similar parts across over-segmented shapes, via feature co-analysis or co-clustering. Learning a set of primitives to represent  shape is another approach to shape decomposition and segmentation, e.g.\ using cuboids~\cite{Tulsiani2017,Sun2019},  superquadrics~\cite{Paschalidou2019}, convex polyhedra~\cite{Deng2020}, or implicit functions~\cite{Genova2020}. Chen \etal~\shortcite{baenet} trained a branched autoencoder network, {Bae-Net} for shape segmentation, in which each branch learns an implicit representation for a meaningful shape part. All the above methods rely on geometric features for segmentation and do not take semantic information into consideration, which may lead to  results inconsistent with human-defined semantics.

\subsection{Supervised 3D segmentation} Various supervised methods perform 3D segmentation using deep neural networks trained on a large number of labeled 3D shapes or scenes~\cite{Guo2019Survey}. Xie \etal~\shortcite{Xie2015} project a 3D shape into multiview images and use 2D CNNs to enhance the segmentation. Kalogerakis \etal~\shortcite{Kalogerakis2017} combine CRF with multiview images to boost  segmentation performance. Dai \etal~\shortcite{dai20183dmv} back-project the feature learned by multiview images to 3D to conduct scene segmentation. Qi \etal\shortcite{Qi2016} use a point-based network to predict per-point semantic labels by combining  global and pointwise features. Other works~\cite{qi2017pointnetplusplus,PointCNN,Thomas2019} enhance  feature propagation by using per-point local information. Wang \etal\shortcite{Wang2019a} and Hanocka \etal\shortcite{meshcnn} build graphs from point sets and conduct message passing on graph edges while further methods~\cite{Kalogerakis2010,GeoCNN,Poulenard2018,PFCNN} directly perform CNN computation on mesh surfaces. Song \etal\shortcite{song2017semantic} conduct  scene semantic segmentation with the help of the scene completion task. For efficiency, many works ~\cite{Wang2017,Graham2018,choy20194d,zhang2020fusion,Huang2021supervoxel} use sparse voxels or supervoxels to reduce the computational and memory costs while achieving better segmentation results. Unlike these supervised methods that require a large amount labeled data, we leverage a few labeled data items and a large amount of unlabeled data for effective segmentation.

\subsection{Weakly-supervised 3D segmentation} 3D shapes in many shape repositories are modeled by artists and often come with rich metadata, like part annotations and part hierarchies. Although part-related information may be inconsistent with the ground truth, it can be used to weakly supervise the training of shape segmentation networks. Yi \etal~\shortcite{Yi2017a} learn hierarchical shape segmentations and labeling from noisy scene graphs from online shape repositories and transfer the learned knowledge to  new geometry. Muralikrishnan \etal~\shortcite{Tags2Parts} discover semantic regions from shape tags. Wang \etal~\shortcite{Wang2018b} learn to group existing fine-grained and meaningful shape segments into semantic parts. Sharma \etal~\shortcite{Sharma2019} embed 3D points into a feature space based on the annotated part tag and group hierarchy and then fine-tune the point features with a few labeled 3D data items for shape segmentation. Zhu \etal~\shortcite{adacoseg2020} utilize part information from a 3D repository to train a part prior network for proposing per-shape parts for an unsegmented shape, then train a co-segmentation network to optimize part labelings across the input dataset.  Xu \etal~\shortcite{Xu2020} learn shape segmentation with an assumption that each shape in the large training dataset has at least one labeled point per semantic part. Unlike these weakly supervised methods, our method  requires no additional weak supervision on unlabeled data.

\subsection{Unsupervised 3D pretraining}

Unsupervised pretraining~\cite{Bengio2013} has demonstrated its advantage in many computer vision and natural language processing tasks, where a feature encoding network is pretrained on a large amount of unlabeled data and then is fine-tuned for downstream tasks using a small amount of labeled data. For 3D analysis tasks, Hassani and Haley~\shortcite{Hassani2019} pretrain a multi-scale graph-based encoder with the ShapeNet dataset~\cite{shapenet2015}  using a multi-task loss. Wang \etal~\shortcite{Wang2020a}  use  multiresolution instance discrimination loss for pre-training, while Hou \etal~\shortcite{Hou2021} and Xie \etal~\shortcite{Xie2020} employ  contrastive loss. Instead of using this two-step training: pretraining and fine-tuning,  our network is trained with both labeled and unlabeled data from the beginning. Given the same amount of labeled data, our semi-supervised method is superior to a fine-tuned pretrained network on 3D object segmentation and indoor scene segmentation.

\subsection{Semi-supervised segmentation}
Semi-supervised learning tries to employ unlabeled data to facilitate supervised learning, thus reducing the amount of labeled data needed for training; see \cite{semisupervisedsurvey} for a detailed survey. Many approaches were first developed for image classification, like temporal ensembling~\cite{laine2016temporal} that aggregates the prediction of multiple previous network evaluations,  Mean-Teacher~\cite{tarvainen2017mean} that averages model weights instead of predictions, FixMatch~\cite{FixMatch} that uses confidence-aware pseudo-labels of weakly-augmented data to guide the strongly-augmented data prediction, and MixMatch~\cite{berthelot2019mixmatch} that guesses low-entropy labels for data-augmented unlabeled data and mixes labeled and unlabeled data using {MixUp}. 
For image segmentation, Ouali \etal~\shortcite{Ouali2020} utilizes \emph{cross-consistency} to train image segmentation networks, where  pixel features extracted by the encoder are perturbed and enforced to be consistent with network predictions after decoding. Ke \etal~\shortcite{Ke2020} use two networks with different initializations and dynamically penalize inconsistent pixel-wise predictions for the same image input. French \etal~\shortcite{French2020} improve image segmentation accuracy by imposing strong augmentation on  unlabeled training images via region masking and replacement. Wang \etal~\shortcite{Wang2021miccai} employ the Mean-Teacher model with improved uncertainty computation and use auxiliary tasks with task-level consistency for medical image segmentation.  Unlike the above semi-supervised image segmentation methods that leverage pixel-level consistency only, or use task-level consistency, our approach utilizes 3D shape part hierarchy and maximizes 3D segmentation consistency at multiple levels, including point-level, part-level and hierarchical level.

For 3D segmentation, {Bae-Net}~\cite{baenet} can learn a branched network from labeled data and unlabeled data for shape segmentation. Although this approach works well for segmenting 3D shapes into a few large parts, it is nontrivial to extend it to many fine-grained semantic 3D segments due to its large network size, and it is unclear whether it can handle the large variety of part structures well.  Wang \etal~\cite{Wang2020b} propose to retrieve a similar 3D shape with part annotations from a mini-pool of shape templates for a given input 3D shape, and learn a transformation to morph the template shape towards the input shape. From the transformed template, a part-specific probability space is learned to predict point part labels, and part consistency within the training batch is utilized. However, its prediction accuracy can be severely affected by the chosen template and the deformation quality.  

\subsection{Structure-aware shape synthesis}
A set of geometric operations has been developed for generating 3D shapes from shape parts, such as part assembly~\cite{Funkhouser2004,Chaudhuri2011,Xie2013}, structural blending~\cite{Alhashim2014}, and set evolution~\cite{Xu2012}. Although these methods are effective for generating high-quality 3D shapes, some  require special pre-processing and interactive editing. Recent methods composite~\cite{Zhu2018, Huang2015,Wu2020} or edit shapes by learning  structural variations within a large set of segmented 3D shapes~\cite{Mo2020}. Another set of methods~\cite{Fu2017,zcm_sfarr_eg13,Guan2020} utilizes the functionality of shape structures to guide 3D shape synthesis. In our work, we develop a simple and automatic part substitution scheme for generating shapes with proper structural and geometric variations from a small number of labeled 3D shapes, whose quality is sufficient to improve network training. We also notice that  recent point cloud augmentation techniques~\cite{chen2020pointmixup,li2020pointaugment,lee2021regularization} that mix points of different shapes randomly to generate more varied shapes can enhance point cloud classification, and can be extended to shape segmentation \cite{zhang2021pointcutmix}. However, random augmentation does not respect shape structure and can lead to limited improvements only, as our experiments show.  Instead, our part substitution scheme enriches  structural variations of the labeled dataset and improves the network performance.

\section{Method Overview} \label{sec:overview}

\subsection{Input and output}
We assume a set of fine-grained 3D semantic part labels probably with structural hierarchy is pre-defined for a 3D shape category. For instance, a coarse level of a chair structure includes the back, the seat, and the support;  the chair support can be decomposed into a finer granularity level which includes vertical legs, horizontal supports, and other small parts. We denote the hierarchical level number by $K$ and the $K$-th level is the finest level.

Our goal is to predict the hierarchical part labels for each point of the input point cloud and determine its shape part structure. The training data includes a small set of labeled point clouds and a large number of unlabeled point clouds. All point clouds are sampled from shapes within the same shape category, so their part structures are implicitly coherent but also with topological and geometric variations.

\subsection{Base network}
Our semi-supervised learning relies on a 3D network that takes a 3D point cloud as input and outputs the point features. Each point feature is transformed to the probability vectors via two fully-connected (FC) layers and a \textsc{Softmax} for determining its segmentation labels at each granularity level. Here we defer the exact choice of our network structure to \cref{sec:net}.

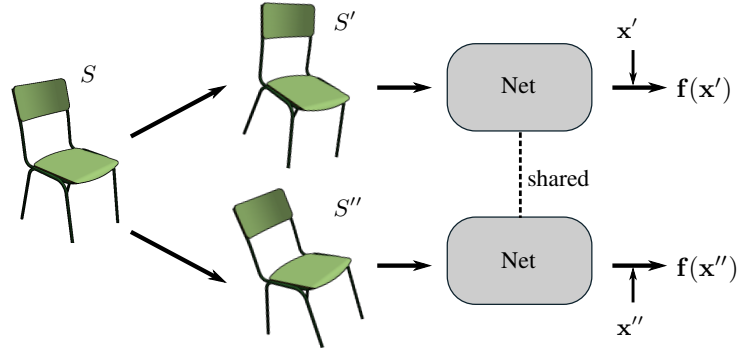
\begin{figure}[t]
\centering
  \begin{overpic}[width=0.6\columnwidth]{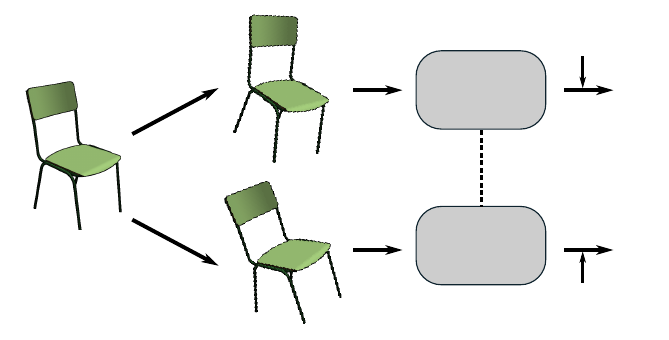}
    \put(13,38){\small $S$}
    \put(47,45){\small $S'$}
    \put(47,20){\small $S''$}
    \put(69.5,36.5){\small Net}
    \put(69.5,13){\small Net}
    \put(73,24){\small shared}
    \put(85,44){\small $\mx'$}
    \put(85,4){\small $\mx''$}
    \put(93,36){$\mf(\mx')$}
    \put(93,12){$\mf(\mx'')$}
   \end{overpic}
  \caption{
    A general neural network setup for 3D semantic segmentation. The network takes a point cloud $S$ as inputs and feeds the two perturbed copies of $S$: $S'$ and $S''$, to the network, separately. The output point features $\mf(\mx)$ of point $\mx$ is transformed to the probability vectors $\mp^{(k)}(\mx_i)$ for determining the segmentation part label at the $k$-th level. The multilevel consistency is built upon the probability vectors of points of $S'$ and $S''$.
  }
  \label{fig:overview} \vspace{-3mm}
\end{figure}

\subsection{Data perturbation for semi-supervised training} For an input point cloud $S$ which is scaled uniformly to fit in a unit sphere, we generate two randomly-perturbed copies of $S$, denoted by $S'$ and $S''$, and pass them to the network during the training stage. In our implementation for shape objects, the perturbation is composed of a uniform scaling with the ratio within interval $[0.75,1.25]$, a random rotation whose pitch, yaw, and roll rotation angles are less than \ang{10}, and random translations along each coordinate axis within the interval $[-0.25,0.25]$. The perturbed point cloud is clipped by the unit box before feeding to the network. This perturbation strategy is the same as the approach of \cite{Wang2020a} for unsupervised pre-training.
The data perturbation makes the trained network more robust and helps build our multilevel consistency between the perturbed shape copies. Without loss of generality, $S'$ and $S''$ are denoted by $\{\mx'_1, \ldots, \mx'_n\}$ and $\{\mx''_1, \ldots, \mx''_n\}$, and $\mx'_i$ and $\mx''_i$ are the two perturbation copies of $\mx_i \in S$. The network with perturbed data is illustrated in \cref{fig:overview}.

\subsection{Notations}
We use the following notations in the paper for convenience.
\begin{description}[leftmargin=0pt,itemsep=1pt]
    \item[$S$] --- The input point cloud: $\{\mx_1,\ldots,\mx_n \in \mathbb{R}^3\}$.
    \item[$L^{(k)} \in \mathbb{N}^+$]  --- The maximal number of semantic labels at the $k$-th level.
    \item[$\mp^{(k)} (\mx_i) \in \mathbb{R}^{L^{(k)}}$] --- Probability vector of $\mx_i$ at the $k$-th level.
    \item[$\mq^{(k)}(\mx_i) \in \mathbb{R}^{L^{(k)}}$] --- One-hot vector of $\mx_i$ at the $k$-th level, corresponding to the ground-truth semantic label of $\mx_i$.
\end{description}

\subsection{Loss design}
For the labeled point cloud $S$, we use the cross-entropy loss to penalize the dissimilarity of point semantic labels of $S'$ and $S''$ with the ground truth in multilevels as follows:
\begin{equation}
    L_{\texttt{seg}}(S',S''):=  \frac{1}{2n} \sum_{i=1}^n \sum_{k=1}^K \bigl[
    g_{\texttt{ce}}\bigl(\mq^{(k)}(\mx_i), \mp^{(k)}(\mx'_i)\bigr)  \bigr.   +  \bigl. g_{\texttt{ce}}\bigl(\mq^{(k)}(\mx_i), \mp^{(k)}(\mx''_i)\bigr)\bigr],
\end{equation}
where $ g_{\texttt{ce}}(\cdot,\cdot)$ is the standard cross-entropy loss.

For both unlabeled and labeled inputs, we introduce the multilevel consistency loss in \cref{sec:consistency} to ensure the network outputs of $S'$ and $S''$ are consistent with each other.

\subsection{Labeled data augmentation}
The structure of labeled 3D shapes offers a great possibility for synthesizing new shapes with semantics. In \cref{sec:aug}, we propose a simple part-substitution method to enrich the labeled shape set, which can improve the performance of both supervised and semi-supervised approaches.

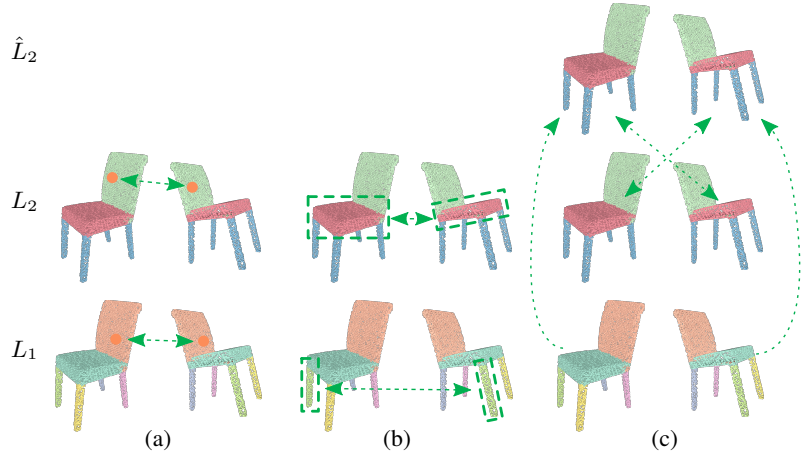
\begin{figure}[t]
\centering
  \begin{overpic}[width=0.6\columnwidth]{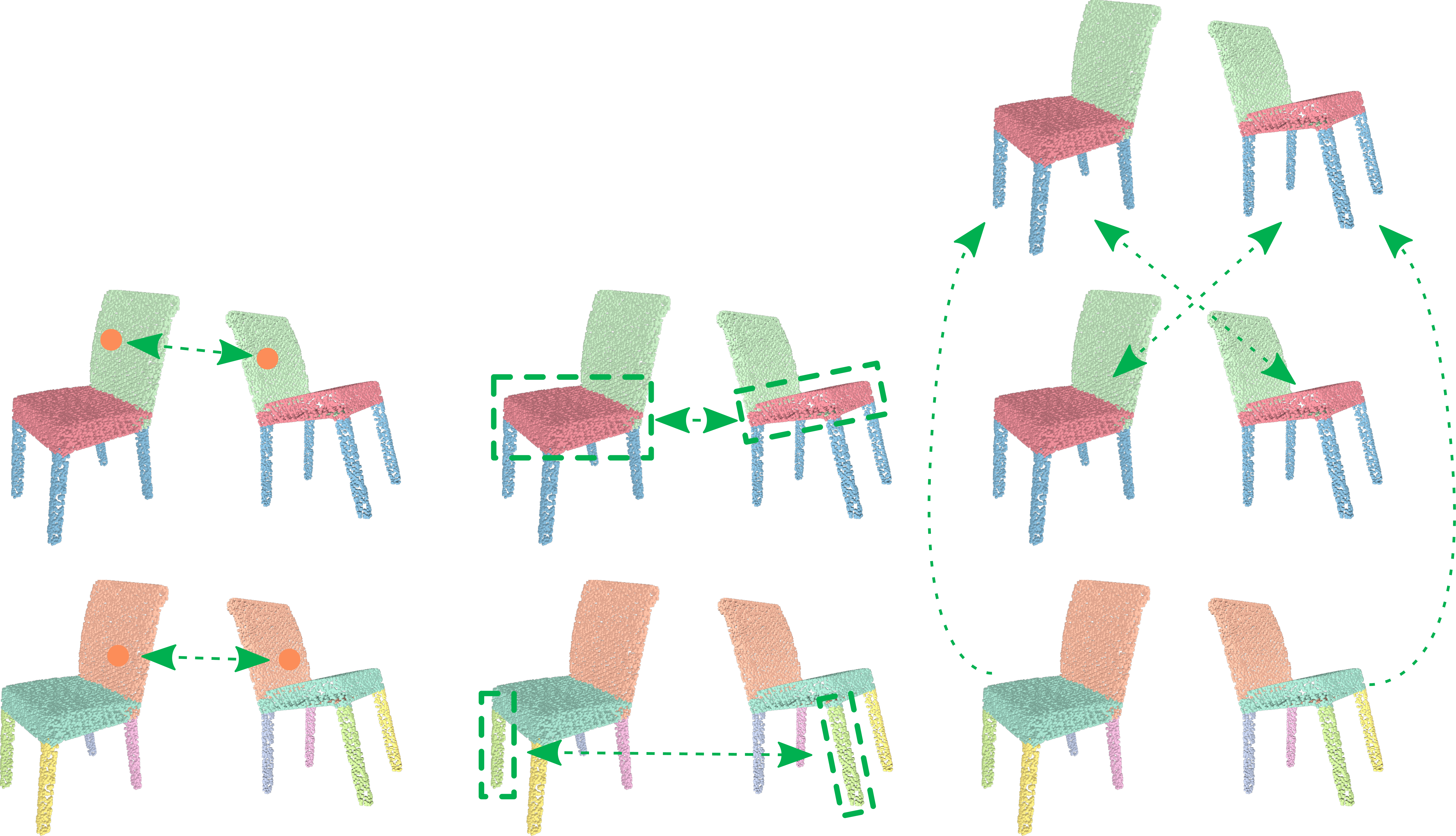}
    \put(12,-2){\small (a)}
    \put(44,-2){\small (b)}
    \put(80,-2){\small (c)}
    \put(-6,10){\small $L_1$}
    \put(-6,30){\small $L_2$}
    \put(-6,50){\small $\hat{L}_2$}
   \end{overpic}
  \caption{
    Multilevel consistency on two perturbed copies of a chair model  having a 2-level hierarchy. $L_1$ and $L_2$ are the fine and coarse levels, respectively. Point color corresponds to predicted part label at each level. The segmentation of $\hat{L}_2$ is the pseudo-part prediction at $L_2$ inferred from $L_1$ according to the predefined shape hierarchy. 
    (a) Point-level consistency built on the corresponding point pairs between two copies at each level. (b) Part-level consistency  built on  parts with the same semantics between two copies at each level. (c) hierarchical consistency built on the corresponding points between the shape copies on  $\hat{L}_2$ and $L_2$.
  }
  \label{fig:consistency} 
  \end{figure}

\section{Multilevel consistency} \label{sec:consistency}
We now introduce our multilevel consistency for utilizing unlabeled data for network training. The multilevel consistency builds on point-level (\cref{subsec:point}), part-level (\cref{subsec:part}), and  hierarchical level (\cref{subsec:hierarchy}) consistency, and is illustrated in \cref{fig:consistency}.

% \input{consistency_figure}
%%%%%%%%%%%%%%%%%%%%%%%%%%%%%%%%%%%%%%%%%%%%%%%%%%%%%%%%%%%%%%%%%%%%%%%%%%%%%%%%%%%
\subsection{Point-level consistency} \label{subsec:point}

A pair of points, $\mx'_i \in S'$ and $\mx''_i \in S''$, should have probability vectors  as similar as possible due to  self-consistency (see \cref{fig:consistency}(a)). Based on this property, we build a point-level consistency loss $L_{\mathrm{point}}$ upon their probability vectors using the symmetric KL-divergence loss $D_{KL}$:
\begin{equation}
    L_{\mathrm{point}} =  \frac{1}{2n}\sum_{i=1}^n \sum_{k=1}^K \left[  D_{KL} \left(\mp^{(k)}(\mx'_i) \parallel  \mp^{(k)}(\mx''_i)\right) \right. + \left. D_{KL} \left(\mp^{(k)}(\mx''_i) \parallel  \mp^{(k)}(\mx'_i)\right) \right].
\end{equation}
Point-level consistency is a simple extension of pixel-level consistency which has been extensively used in semi-supervised image segmentation.
The KL-divergence loss can be replaced with the MSE loss; the latter has a better performance on semi-supervised image classification in \cite{laine2016temporal, tarvainen2017mean}. However, we found that they have similar performance on 3D segmentation, and indeed the former is slightly better.

%%%%%%%%%%%%%%%%%%%%%%%%%%%%%%%%%%%%%%%%%%%%%%%%%%%%%%%%%%%%%%%%%%%%%%%%%%%%%%%%%%%
\subsection{Part-level consistency} \label{subsec:part}
Due to data perturbation, the predicted part distributions of $S'$ and $S''$ at the same part level can be different. We impose a novel part-level consistency to minimize this difference.

For a point $\mx' \in S'$, its predicted part label at the $k$-th level is determined by $\argmax_{m} \{p^{(k)}_m(\mx'_i), m=1,\ldots,L^{(k)}\} $, where $p^{(k)}_m(\mx'_i)$ is the $m$-th component of $\mp^{(k)}(\mx'_i)$. Using  the predicted part labels of all points at the $k$-th level, we can partition $S'$ into a set of parts, denoted  $\{\mP^{(k)}_1,\cdots, \mP^{(k)}_{L^{(k)}}\}$, where some sub-partitions can be empty. We call these parts a \emph{pseudo-partition}. On $S''$, we also compute a pseudo-partition, denoted  $\{\mQ^{(k)}_1,\cdots, \mQ^{(k)}_{L^{(k)}}\}$.

For a pseudo-part $\mP^{(k)}_l$, we define two statistical quantities: \emph{belonging-confidence} and \emph{outlier-confidence}, denoted by $\BC(\mP^{(k)}_l)$ and $\OC(\mP^{(k)}_l)$, respectively.  The belonging-confidence measures the confidence with which  points in $\mP^{(k)}_l$ belong to $\mP^{(k)}_l$ and the outlier-confidence measures the confidence with which the remaining points outside $\mP^{(k)}_l$ do not belong to  $\mP^{(k)}_l$. They are defined as follows.
\begin{equation}
    \begin{aligned}
        \BC(\mP^{(k)}_l, S') = & \Mean\{ p^{(k)}_l(\my), \; \forall \my \in S' \cap \mP_l^{(k)}\};      \\
        \OC(\mP^{(k)}_l, S') = & \Mean\{ p^{(k)}_l(\my), \; \forall \my \in S' \backslash\mP_l^{(k)}\}.
    \end{aligned}
\end{equation}

As the pseudo-partitions of $S'$ and $S''$ should be consistent with each other, we can impose the pseudo-partition of $S'$ onto $S''$, \ie partition  $S''$ according to the point assignment of $\{\mP^{(k)}_1,\cdots, \mP^{(k)}_{L^{(k)}}\}$, and compute the corresponding belonging-confidence and the outlier-confidence values on $S''$. Because of  self-consistency, we expect these values to be as close as possible to the corresponding values computed on $S'$. Similarly, we can also impose the pseudo-partition of $S''$ onto $S'$ in a similar way. We call this type of consistency \emph{part-level consistency}, and define the loss function as follows:
\begin{equation}
    \begin{aligned}
        L_{\mathrm{part}}  =  \sum_{k=1}^K \sum_{j=1}^{L^{(k)}}   \left[ \right.& \alpha \|\BC(\mP^{(k)}_j, S') - \BC(\mP^{(k)}_j, S'')\|^2 + \beta \|\OC(\mP^{(k)}_j, S') - \OC(\mP^{(k)}_j, S'')\|^2 +
        \\   &\alpha \|\BC(\mQ^{(k)}_j, S'') - \BC(\mQ^{(k)}_j, S')\|^2 + \beta \|\OC(\mQ^{(k)}_j, S'') - \OC(\mQ^{(k)}_j, S')\|^2 \left.\right].
    \end{aligned}
\end{equation}
Here $\alpha$ and $\beta$ are dynamically adjusted: $\alpha = \beta = 1/2$ when the sub-partition appearing in the $\BC$ term is nonempty, otherwise we set $\alpha = 0$, $\beta =1$.  \cref{fig:consistency}(b) illustrates part consistency on a chair model.

%%%%%%%%%%%%%%%%%%%%%%%%%%%%%%%%%%%%%%%%%%%%%%%%%%%%%%%%%%%%%%%%%%%%%%%%%%%%%%%%%%%
\subsection{Hierarchical consistency} \label{subsec:hierarchy}
For a shape category possessing a part structure hierarchy, the semantic segmentation labels at different levels are strongly correlated. We propose \emph{hierarchical consistency} to utilize this structure prior.

For a point $\mx \in S$, we can use its probability vector at level $(k+1)$ to infer its part label probability at level $k$, \ie its parent level,  just by merging the probability values of $\mp^{(k+1)}(\mx)$ to form the probability vector at level $k$, according to the predefined shape hierarchy. For instance, in the chair structure, suppose the chair arm contains two parts: a vertical bar and a horizontal bar. We add the probability values of the vertical bar and horizontal bar together and set their sum as the probability value of the chair arm.

In this way, we can create a \emph{pseudo-probability vector} for $\mx$ at level $k$, denoted  $\hat{\mp}^{(k)}(\mx)$. Ideally $\hat{\mp}^{(k)}(\mx'_i)$ should be the same as $\mp^{(k)}(\mx''_i)$ predicted by the network and vice versa.
We call this relation \emph{hierarchical consistency}, and define a loss function on the points of $S'$ and $S''$ using KL-divergence as follows:
\begin{equation} \label{eq:seg}
    L_{\mathrm{h}} :=  \frac{1}{2n}\sum_{i=1}^n \sum_{k=1}^{K-1} \left[  D_{KL}\left( \hat{\mp}^{(k)}(\mx'_i) \parallel \mp^{(k)}(\mx''_i) \right) \right.+ \left.D_{KL}\left( \hat{\mp}^{(k)}(\mx''_i) \parallel \mp^{(k)}(\mx'_i) \right) \right].
\end{equation}
\cref{fig:consistency}(c) illustrates  hierarchical consistency on a chair model.

Note that the above hierarchical consistency is defined across two perturbed shapes. In fact, it is possible to impose  hierarchical consistency on a single perturbed shape using $D_{KL}\bigl( \hat{\mp}^{(k)}(\mx'_i) \parallel \mp^{(k)}(\mx'_i) \bigr)$, but in practice we find that these consistency terms are easily satisfied as the multilevel probability vectors of the same shape are highly correlated, so  do not give much assistance in semi-supervised training.

\section{Multilevel part substitution} \label{sec:aug}

\begin{figure}[t]
    \centering
\scalebox{1.17}{
    \begin{overpic}[width=0.5\columnwidth]{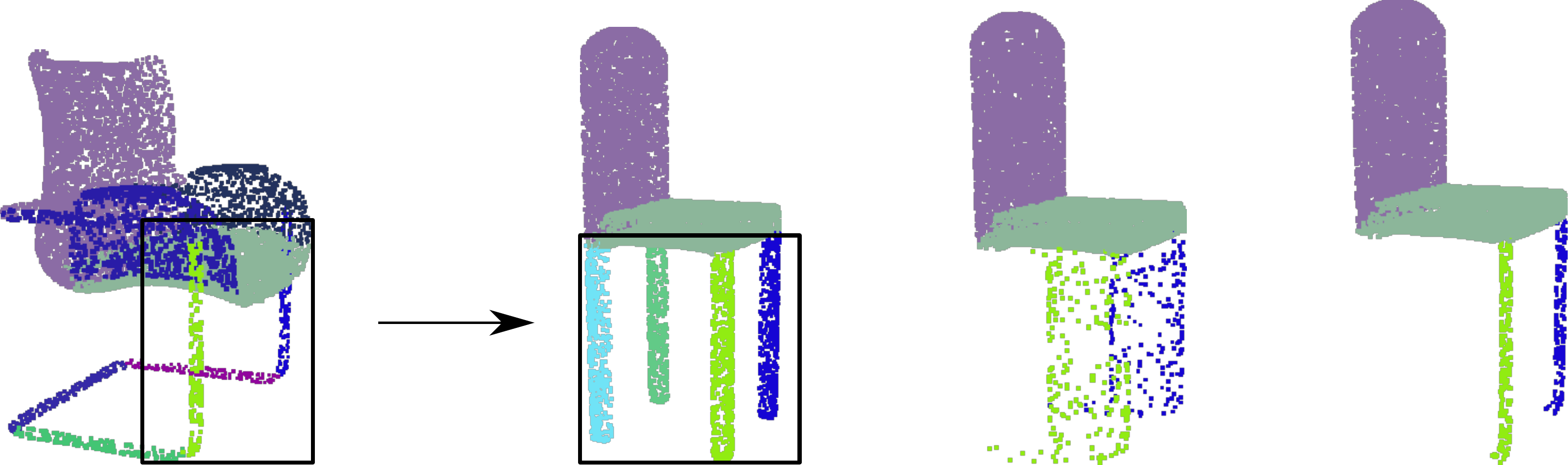}
        % \put(9,-5){\small \bfseries (a)}
        \put(9,-7){\small (a)}
        \put(41,-7){\small (b)}
        \put(67,-7){\small (c)}
        \put(91,-7){\small (d)}
    \end{overpic}
    }
        \vspace{7mm}
    \caption{Multilevel part substitution.  The two front legs of the chair in (a) are selected to replace the four legs in (b). (c): unsatisfactory substitution by aligning the two regions directly. (d): good substitution by aligning the common parts (front legs) first. While the result is not physically plausible, it is suitable for training.}
    \label{fig:strange} 
    \end{figure}
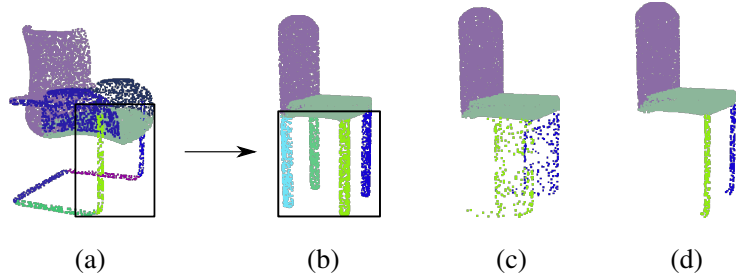

We propose a simple multilevel part-substitution algorithm to enrich the labeled 3D shapes for training. Given a randomly sampled labeled shape $S$, our algorithm executes the following steps to synthesize new shapes with geometry and structural variation.

Part selection is carried out first.  We treat the  hierarchical structure of shape $S$ as a tree, where each shape part is a tree node. We visit each node from the coarsest level to the finest level.  For a node at level $k$,  a uniform random number in $[0,1]$ is generated. If the number is smaller than a predefined threshold $\theta_k$, we set the subtree under this node as a replacement candidate and stop visiting its children.  Finally, we collect a set of subtrees to be replaced.

Next, part substitution is performed. For a part subtree $P$ in the candidate list, we randomly select a subtree $Q$ from those other shapes in which the root node of $Q$ has the same semantic label as $P$'s root node. Note that simple substitution of $P$ by $Q$ may result in strange-looking and partly-overlapping results (see \cref{fig:strange}(c)), so we replace $P$ by $Q$ as follows, to avoid unpleasant results as much as possible. We consider two cases.
 
 If the leaf nodes of $P$ and $Q$ have no common parts sharing the same semantics,  we simply compute the affine transformation from the bounding box of $Q$ to the bounding box of $P$ and apply it to $Q$ when replacing $P$.

However, if $P$ and $Q$ share some common semantic parts, denoted  $P_s \in P$, $Q_s \in Q$, we align $Q_s$ and $P_s$ first to avoid odd results.          The alignment transformation matrix is applied to $Q$ directly. We also rescale the transformed $Q$ to ensure that it is inside the original bounding box of $S$, to make the result visually plausible: see \cref{fig:strange}(d).

\begin{figure}[t]
    \centering
    \includegraphics[width=0.7\linewidth]{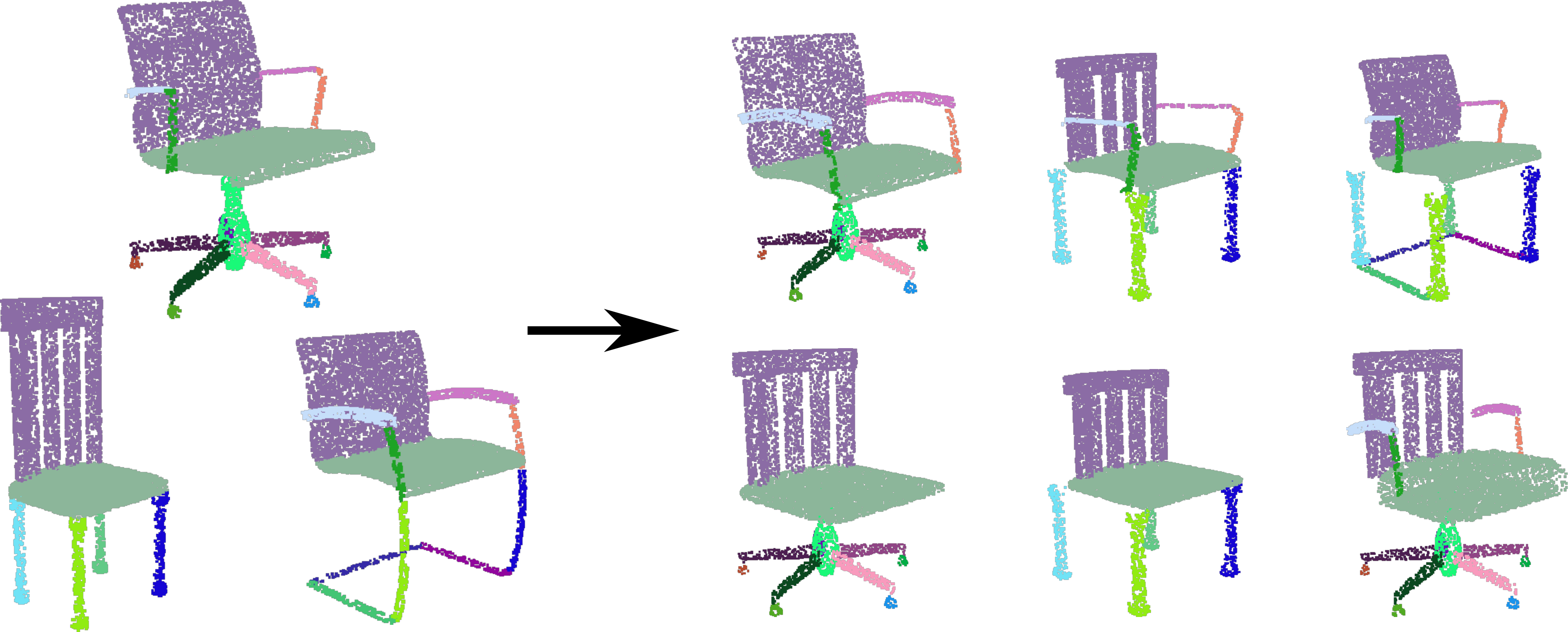}
    \caption{Randomly generated shapes from three labeled chairs. Point colors correspond to  semantic IDs at the finest level.}
    \label{fig:gen} 
\end{figure}

The $\theta_k$ values affect the degree of structure variation: frequent substitutions at the coarse level bring more structural variations. In our experiments, we set all $\theta_k$s to $0.5$. \cref{fig:gen} shows a set of novel chairs synthesized from three chairs. More synthesized shapes used in our experiments are illustrated in Appendix C.

After generation, as all parts of the synthesized shapes inherit their original semantics, these shapes can be used as labeled data. Note that after part substitution, two different shape parts in a shape may  overlap. We detect  points inside these overlapping regions using a simple nearest neighbor search and do not use their labels during training, to avoid contradictions.

\section{Network Design, Loss and Training Data} \label{sec:net}
In this section, we present details of our network structure, loss function and training batches. %and training protocol.

\subsection{Network structure}
\begin{figure}[t]
    \centering
    \includegraphics[width=0.6\columnwidth]{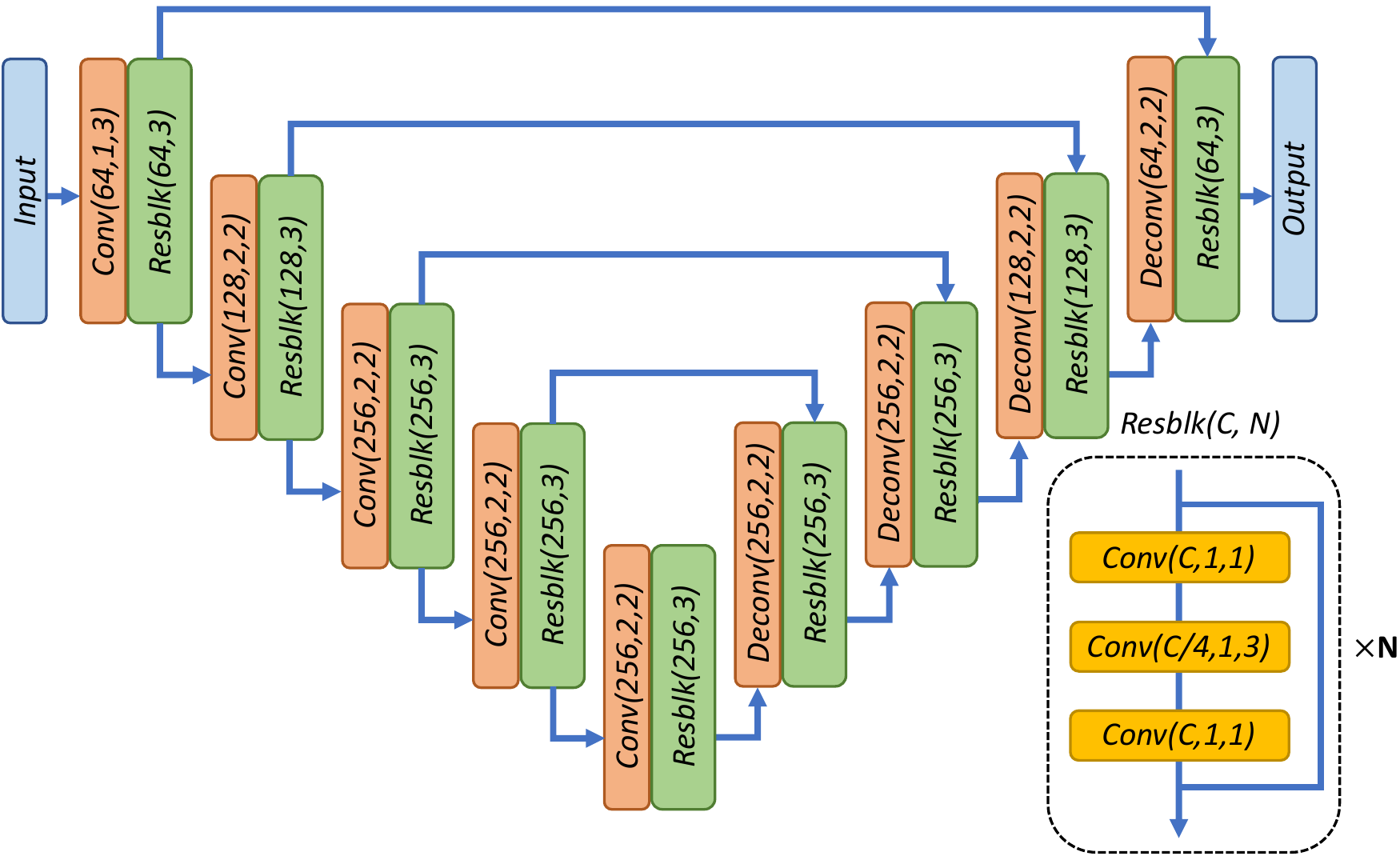}
    \caption{Octree-based U-Net structure for shape segmentation. Conv(C, S, K) and Deconv(C, S, K) represent octree-based convolution and deconvolution. C, S, K are the number of output channels, stride, and kernel size.  The network structure for indoor scene segmentation is provided in the appendix.}
    \label{fig:unet} 
\end{figure}

We use an octree-based U-Net structure as our base network. The network is built upon the efficient open-source octree-based CNN~\cite{Wang2017,Wang2020}.  The U-Net structure has five and four levels of domain resolution, as illustrated in \cref{fig:unet} and Appendix A, for object segmentation and scene segmentation, respectively. The maximum depths of the octree for 3D object segmentation and scene segmentation are $6$ and $9$, respectively. The input point cloud is converted to an octree first, whose nonempty finest octants store the average of the normals of the points within them. The point feature for a given point is found by trilinear interpolation within the octree.
The numbers of network parameters for 3D object segmentation and 3D scene segmentation are $5.3\times10^6$ and $39.2\times10^6$, respectively. We call our network {MCNet}, for multi-consistency 3D deep learning network. In \cref{sec:result} we also demonstrate the efficiency of our approach based on other point-based backbones.

\subsection{Loss function} \label{subsec:loss}
Given a point cloud $S$ in a training batch, the loss defined on its two randomly-perturbed copies $S'$ and $S''$ is:
\begin{equation}
    \begin{aligned}
        L_{\mathrm{tc}} =   \gamma L_{\mathrm{seg}}(S',S'') + \lambda_\mathrm{pts} L_{\mathrm{point}} + \lambda_\mathrm{part} L_{\mathrm{part}} +\lambda_{h}  L_{\mathrm{h}};
    \end{aligned}
\end{equation}
$\gamma = 0$ if $S$ is an unlabeled point cloud.

\subsection{Training batch construction} 
Half of the batch data is randomly selected from the labeled data, and the rest is randomly selected from the unlabeled dataset. If synthetic labeled data (\cref{sec:aug}) are available, half of the labeled data in the batch is selected from them, and the remainder is selected from the original labeled data. The labeled data in a batch may be duplicated if the labeled dataset is quite small. The network is trained from scratch with random initialization.

\section{Experiments and Analysis} \label{sec:result}
In this section, we demonstrate the efficacy and superiority of our semi-supervised approach on shape segmentation  and scene segmentation, and an ablation study to validate our design.

Our experiments were conducted on a Linux server with a \SI{3.6}{GHz} Intel Core I7-6850K CPU  and a Tesla V100 GPU  with \SI{16}{GB} memory for experiments on shape objects, and a Tesla V100  with \SI{32}{GB} memory for  indoor scenes. We implemented our network using the TensorFlow framework~\cite{tensorflow2015-whitepaper}.

\subsection{Shape segmentation} \label{subsec:shapeseg}

\subsubsection{Datasets}
Our semi-supervised 3D segmentation approach was evaluated on the following datasets with different ratios of labeled data.

\begin{figure}[t]
    \centering
\scalebox{1.1}{    \begin{overpic}[width=0.6\columnwidth]{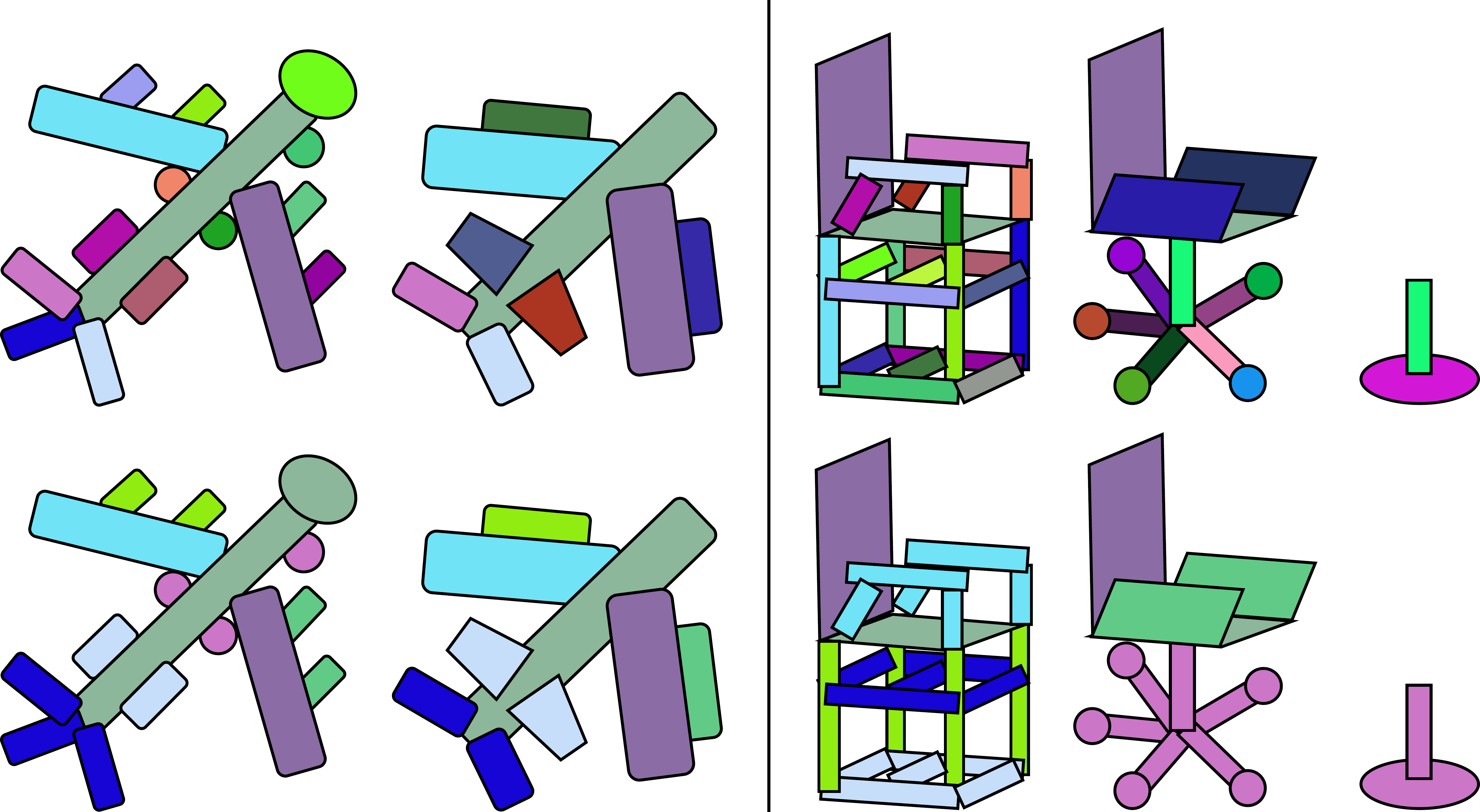}
    \put(16.5,-4){\small Airplane}
    \put(74,-4){\small Chair2}
    \put(87,48){\small fine level}
    \put(85,22){\small coarse level}
    \end{overpic}
    }
    \vspace{3mm}
    \caption{The two-level fine-grained and hierarchical structures for our Airplane and Chair2 datasets.  Unique colors at the fine level correspond to  distinguishable shape parts, as a segmentation label.  Several parts at the fine level are merged to form a unique segmentation label at the coarse level, and assigned the same color. Both Chair2 and Airplane have 8 different part labels at the coarse level, while at fine levels they have 36 and 20 part labels, respectively. }
    \label{fig:hierarchy} 
\end{figure}
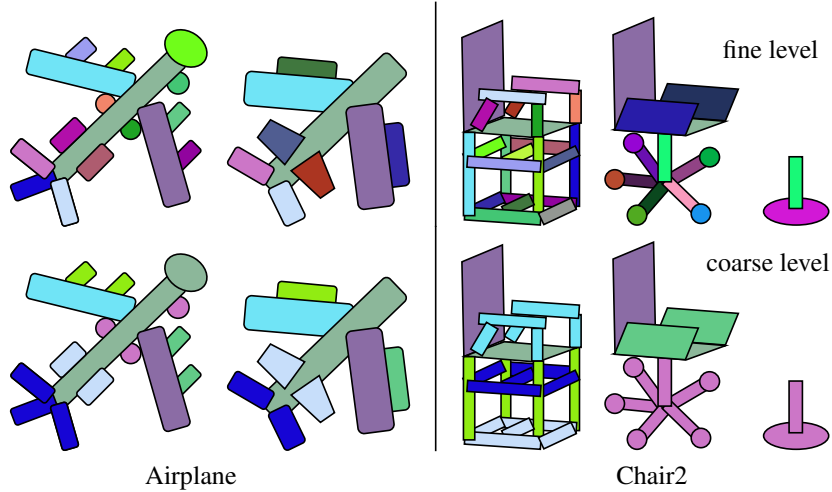

\emph{PartNet}. The PartNet dataset~\cite{Mo2019} provides fine-grained, hierarchical segmentation of 26671 models in 24 object categories, and defines three levels (coarse, medium, fine) of segmentation for the benchmark.

\emph{Shape categories with customized hierarchy}.
We defined a two-level part hierarchy on two shape categories: the Chair from PartNet and the Airplane from ShapeNet~\cite{shapenet2015}, to further validate the effectiveness of our approach on other shape data and structural hierarchies. At the fine level, our new data provides finer-grained part labels than PartNet. For instance, each chair leg is treated as a different object part while all legs of a chair belong to a single part in the PartNet level-3 segmentation. The hierarchical relationship between each level is also differs from PartNet.
The hierarchical fine-grained structures of these two categories are illustrated in \cref{fig:hierarchy}. To avoid confusion, we call our chair dataset Chair2.  The Chair2 dataset contains 3303 models for training and 826 models for testing, and the Airplane dataset contains 1404 models for training and 366 models for testing.

\emph{ShapeNetPart}.   ShapeNetPart \cite{Yi2017a} contains 16 shape categories from ShapeNet. Each model is a point cloud with $2$--$6$ part labels without a structural hierarchy.

\emph{ScanNet}. The ScanNet dataset~\cite{dai2017scannet} contains 1613 3D indoor scenes with 20 labels for semantic segmentation. The numbers of scenes for  training, validation, and testing are 1201, 312, and 100, respectively.

For the above datasets, we used a fixed seed to randomly pick a small fraction of the labeled training data, around 2\%, and set it as the labeled data for semi-supervised training, and the remaining labeled training data was treated as unlabeled data in our training: no label information was utilized during  training and part substitution. The original testing dataset was used as unseen test data for evaluating the trained network.

Each training batch contained 16 shapes. A maximum of 80000 iterations was used. We used the SGD optimizer with a learning rate of {0.1}, decayed by a factor of {0.1}  at the 40000-th  and 60000-th iterations.  For the loss function, we empirically set  $\lambda_\mathrm{pts}=\lambda_\mathrm{part}=\lambda_{h}=0.01$, via a simple grid search on the four biggest categories of PartNet. To conduct a statically meaningful evaluation,  we ran training on each shape category three times with different randomly-selected labeled data, and report average results.

\begin{table*}[t]
    \centering
    \caption{Segmentation results on PartNet. All metrics are averaged across 24 categories. \textbf{r} is the fraction of labeled data used for  training. }
    \scalebox{1}{%\scriptsize
        \begin{tabular}{cc|cc|cc|cc|cc}
            \toprule
                                    &                        & \multicolumn{2}{c}{\thead{Coarse Level}} & \multicolumn{2}{c}{\thead{Medium Level}} & \multicolumn{2}{c}{\thead{Fine Level}} & \multicolumn{2}{c}{\thead{Avg}} \\
            \midrule
            \thead{r}               & \thead{Method}         & \thead{p-mIoU}         & \thead{s-mIoU} & \thead{p-mIoU}         & \thead{s-mIoU} &  \thead{p-mIoU}         & \thead{s-mIoU} & \thead{p-mIoU}         & \thead{s-mIoU} \\
            \midrule
            \multirow{2}{*}{$2\%$}  & {MIDNet}                 & 44.7 & 63.5 & 29.8 & 43.9 & 24.9 & 40.6 & 38.2 & 54.7   \\
                                    & {MCNet}                  & \textbf{47.6} & \textbf{68.4} & \textbf{33.4} & \textbf{48.9} & \textbf{27.4} & \textbf{45.8} & \textbf{41.3} & \textbf{60.0}  \\
            \midrule
            \multirow{2}{*}{$5\%$}  & {MIDNet }                & 53.2 & 69.1 & 35.9 & 48.4 & 32.7 & 46.0 & 46.7 & 60.7   \\
                                    & {MCNet}                  & \textbf{54.9} & \textbf{71.9} & \textbf{38.5} & \textbf{52.0} & \textbf{34.2} & \textbf{49.3} & \textbf{48.8} & \textbf{63.5}  \\
            \midrule
            \multirow{2}{*}{$10\%$} & {MIDNet}                 & 57.5 & 71.7 & 39.7 & 51.9 & 37.6 & 50.3 & 51.6 & 63.8   \\
                                    & {MCNet}                  & \textbf{60.9} & \textbf{75.5} & \textbf{43.9} & \textbf{54.6} & \textbf{40.2} & \textbf{52.0} & \textbf{54.8} & \textbf{67.0}  \\
            \midrule
            \multirow{2}{*}{$20\%$} & {MIDNet}                 & 64.2 & 75.7 & 44.6 & 55.4 & 43.3 & 54.2 & 57.7 & 68.0   \\
                                    & {MCNet}                  & \textbf{65.2} & \textbf{78.5} & \textbf{48.7} & \textbf{58.6} & \textbf{45.2} & \textbf{56.4} & \textbf{59.4} & \textbf{70.7}  \\
            \bottomrule
        \end{tabular}
    }    \label{tab:partnet-eval}
\end{table*}

%%%%%%%%%%%%%%%%%%%%%%%%%%%%%%%%%%%%%%%%%%%%%%%%%%%%%%%%
\subsubsection{PartNet segmentation} \label{subsubsec:partnet}

On all 24 shape categories of PartNet, we experimented with our semi-supervised training scheme with different ratios of labeled data for our {MCNet}: 2\%, 5\%, 10\%, and 20\%. We generated as many randomly synthesized labeled shapes by  part substitution  as the number of  original training data shapes. Following \cite{Mo2019}, we used the following metrics to evaluate the results.

\emph{p-mIoU}. The IoU between the predicted point set and the ground-truth point set for each semantic part category is first computed over the test shapes, then the per-part-category IoUs are averaged. This metric helps evaluate how an algorithm performs for any given part category~\cite{Mo2019}, but does not characterize the segmentation quality at the object level.

\emph{s-mIoU} The part-wise IoU is first computed for each shape, then the mean IoU over all parts is computed on this shape, and finally, these mean IoUs are averaged over all test shapes. This metric is sensitive to  missing ground-truth parts and the appearance of unwanted predicted parts in a shape.

We  choose {MIDNet}~\cite{Wang2020a} as a basis for comparison, which has unsupervised 3D pretraining with a  fine tuning method and provides state-of-the-art results on PartNet with a small amount of labeled data. {MIDNet} was pretrained on the ShapeNet dataset. We fine tuned {MIDNet} with the same limited labeled data as our method, using the multilevel segmentation loss in \cref{eq:seg}.  The results are reported in \cref{tab:partnet-eval}.  Our {MCNet} achieved superior results to {MIDNet} on all  tests at all segmentation levels.

\begin{figure*}[!ht]
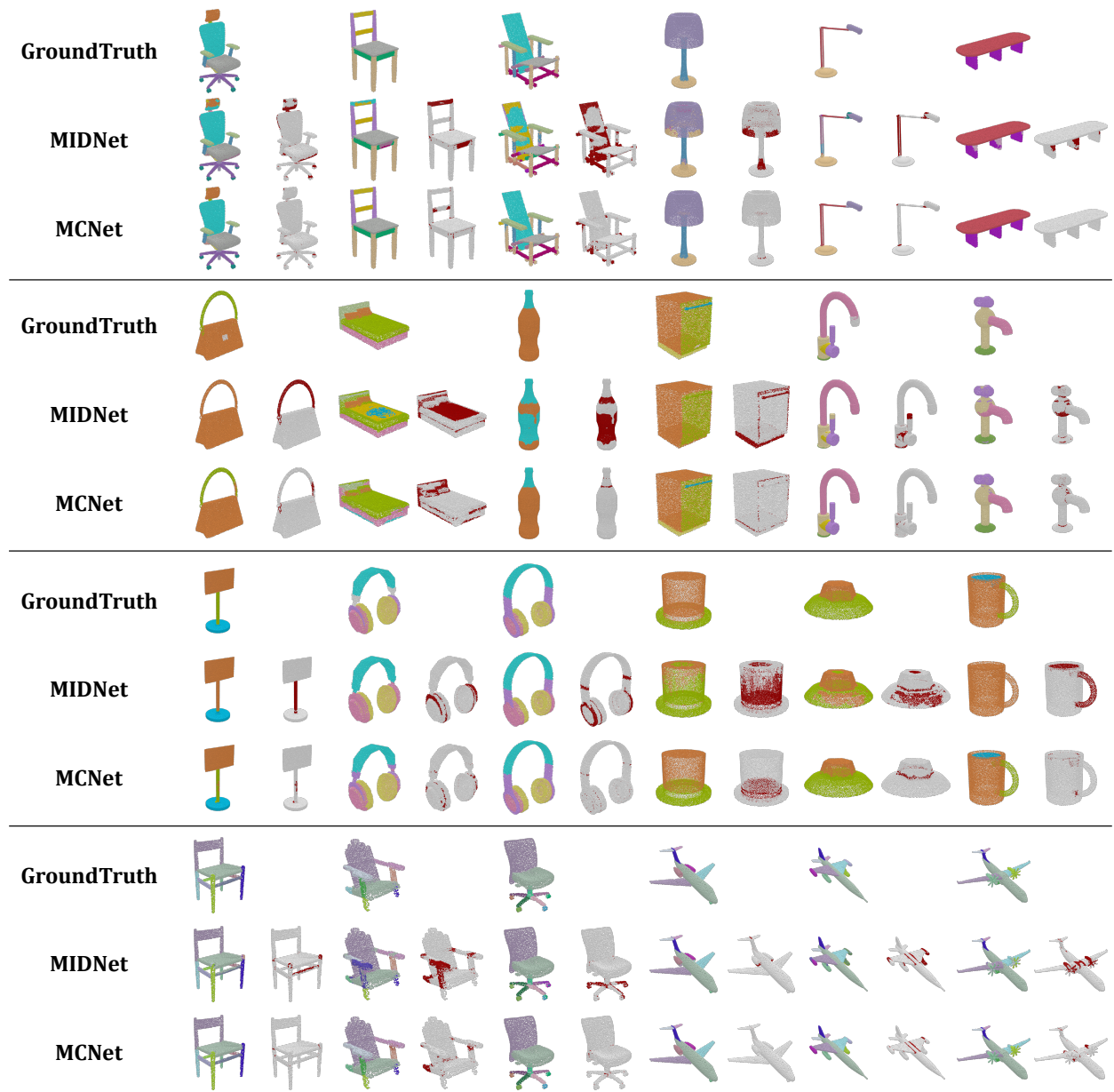

    \centering
    \begin{overpic}[width=\linewidth]{comparison}
    \end{overpic}
    \caption{Fine-level segmentation results from our {MCNet} and {MIDNet}. 2\% labeled data were used in  training.   Point colors in the segmentation results correspond to part ID. In the error maps alongside the segmentation results, red points indicate wrongly predicted labels. The top three sets of examples came from the PartNet test set; those at the bottom came from the Chair2 and Airplane test data.}
    \label{fig:comp}
\end{figure*}

In \cref{fig:comp}, we illustrate  segmentation results and error maps resulting from our approach and {MIDNet} on a set of test shapes, trained with 2\% labeled data. The results clearly show that our method has lower segmentation error.

\begin{table*}[t]
    \centering
    \caption{Segmentation results on PartNet with different backbone networks. All metrics are averaged across 3 levels of the test dataset for 24 categories. \textbf{r} is the proportion of labeled data used for  training. Baseline is the supervised approach with  multilevel segmentation loss. {Ours} is the backbone with our semi-supervised approach.}
    %\resizebox{\columnwidth}{!}{%\scriptsize
        \begin{tabular}{cc|cc|cc|cc}
            \toprule
                                    &                   & \multicolumn{2}{c}{\thead{PointNet++}} & \multicolumn{2}{c}{\thead{PointCNN}} & \multicolumn{2}{c}{\thead{OCNN}}                                                    \\
            \midrule
            \thead{r}               & \thead{Method}    & \thead{p-mIoU}                         & \thead{s-mIoU}                       & \thead{p-mIoU}                   & \thead{s-mIoU} & \thead{p-mIoU} & \thead{s-mIoU} \\
            \midrule
            \multirow{2}{*}{$2\%$}  & {Baseline} & 34.6                                   & 47.6                                 & 35.3                             & 50.6           & 35.8           & 51.2           \\
                                    & {Ours}      & 39.7                                   & 57.7                                 & 40.7                             & 58.3           & 41.3           & 60.0           \\
            \midrule
            \multirow{2}{*}{$5\%$}  & {Baseline} & 42.3                                   & 53.6                                 & 42.9                             & 55.2           & 42.8           & 54.7           \\
                                    & {Ours}      & 47.8                                   & 62.6                                 & 49.2                             & 62.2           & 48.8           & 63.5           \\
            \midrule
            \multirow{2}{*}{$10\%$} & {Baseline} & 48.0                                   & 59.2                                 & 49.1                             & 61.0           & 49.4           & 60.9           \\
                                    & {Ours}      & 51.5                                   & 65.1                                 & 52.3                             & 65.7           & 54.8           & 67.0           \\
            \midrule
            \multirow{2}{*}{$20\%$} & {Baseline} & 53.5                                   & 63.9                                 & 54.0                             & 63.8           & 56.2           & 66.4           \\
                                    & {Ours}      & 56.2                                   & 68.6                                 & 56.5                             & 68.3           & 59.4           & 70.7           \\
            \bottomrule
        \end{tabular}
    \label{tab:backbone-eval}
\end{table*}

We also replaced our octree-based CNN backbone with other popular point-based deep learning frameworks: PointNet++~\cite{qi2017pointnetplusplus} and PointCNN~\cite{PointCNN}, and tested their segmentation performance. \cref{tab:backbone-eval} reports the significant improvements brought by our multilevel consistency and part substitution, compared to their purely-supervised {baseline}. We also found that these backbones did not yield better results than the octree-based CNN backbone.

%%%%%%%%%%%%%%%%%%%%%%%%%%%%%%%%%%%%%%%%%%%%%%%%%%%%%%%%
\subsubsection{Segmentation on shape categories with customized hierarchy}\label{subsubsec:partnet2}
Like the experiments on PartNet, the experiments on Chair2 and Airplane also showed that our approach is significantly better than {MIDNet} (see \cref{tab:partnet2-eval}). Several segmentation results are also illustrated in \cref{fig:comp}.

\begin{table*}[t!]
    \centering
    \caption{Segmentation results on the test dataset containing Chair2 and Airplane. \textbf{r} is the proportion of labeled data used for  training.}
        \begin{tabular}{cc|cc|cc}
            \toprule
                                    &                        & \multicolumn{2}{c}{\thead{Fine Level}} & \multicolumn{2}{c}{\thead{Coarse Level}}                                                                                                                                    \\
            \midrule
            \thead{r}               & \thead{Method}         & \thead{p-mIoU}         &  \thead{s-mIoU}   &    \thead{p-mIoU} &  \thead{s-mIoU} \\
            \midrule
            \multirow{2}{*}{$2\%$}  & {MIDNet}                 & 75.7                & 85.6                  & 87.9      & 87.4  \\
                                    & {MCNet}         & \textbf{82.4}                & \textbf{89.0}                  & \textbf{91.4}      & \textbf{91.4}  \\
            \midrule
            \multirow{2}{*}{$5\%$}  & {MIDNet}                 & 81.0                & 87.5                  & 90.1      & 89.0  \\
                                    & {MCNet}         & \textbf{84.5}                & \textbf{90.2}                  & \textbf{92.1}      & \textbf{92.1}  \\
            \midrule
            \multirow{2}{*}{$10\%$} & {MIDNet}                 & 82.8                & 88.0                  & 90.4      & 89.9  \\
                                    & {MCNet}         & \textbf{85.7}                & \textbf{90.5}                  & \textbf{92.3}      & \textbf{92.2}  \\
            \midrule
            \multirow{2}{*}{$20\%$} & {MIDNet}                 & 83.5                & 89.4                  & 90.9      & 90.8  \\
                                    & {MCNet}         & \textbf{86.0}                & \textbf{91.1}                  & \textbf{92.2}      & \textbf{92.7}  \\

            \bottomrule
        \end{tabular}
    \label{tab:partnet2-eval}
    %\vspace{-3mm}
\end{table*}

\begin{table}[t]
    \centering
    \caption{Segmentation results for different methods on  ShapeNetPart, with 5\% labeled training data.}
%    \scalebox{1}{
%        \scriptsize
        \begin{tabular}{lcc}
            \toprule
            \thead{Method}  & \thead{c-mIoU} & \thead{i-mIoU}\\
            \midrule
            {SO-Net}~\cite{li2018so} & -     & 69.0      \\
            {PointCapsNet}~\cite{Zhao_2019_CVPR} & -     & 70.0      \\
            {MortonNet}~\cite{Thabet_2020_CVPR_Workshops} & -     & 77.1      \\
            {JointSSL}~\cite{JointSSL} & -     & 77.4      \\
            {Multi-task}~\cite{Hassani2019} & 72.1     & 77.7      \\
            {ACD}~\cite{ACD}        & -   & 79.7        \\
            % {SurFit}~\cite{sharma2021surfit}     & 79.0 & 78.7    \\ % the number is not reasonable
            {MIDNet}~\cite{Wang2020a}     & 77.7   &  80.7  \\
            {MCNet}      & \textbf{79.8}  & \textbf{82.2}  \\
            \bottomrule
        \end{tabular}
%    }\vspace{2mm}
    \label{tab:shapenetpart-comp}
%    \vspace{-3mm}
\end{table}

\begin{table}[b]
    \centering
    \caption{Segmentation results for ShapeNetPart. Higher mIoU values are better. \textbf{r} is the proportion of labeled data used for  training.}
%    \scalebox{1}{
%        \scriptsize
        \begin{tabular}{rl|cc}
            \toprule
            \thead{r}               & \thead{Method}  & \thead{c-mIoU} & \thead{i-mIoU} \\
            \midrule
            \multirow{2}{*}{$2\%$}  & {MIDNet} & 73.9           & 78.4           \\
                                    & {MCNet}  & \textbf{76.1}  & \textbf{81.2}  \\
            \midrule
            \multirow{2}{*}{$5\%$}  & {MIDNet} & 77.7           & 80.7           \\
                                    & {MCNet}  & \textbf{79.8}  & \textbf{82.2}  \\
            \midrule
            \multirow{2}{*}{$10\%$} & {MIDNet} & 79.2           & 82.3           \\
                                    & {MCNet}  & \textbf{81.8}  & \textbf{84.2}  \\
            \midrule
            \multirow{2}{*}{$20\%$} & {MIDNet} & 81.7           & 83.1           \\
                                    & {MCNet}  & \textbf{83.0}  & \textbf{84.3}  \\
            \bottomrule
        \end{tabular}
%    }\vspace{2mm}
    \label{tab:shapenetpart-eval}
%    \vspace{-3mm}
\end{table}

\begin{table}[t]
    \centering
    \caption{mIoU results for our method and that of~\cite{Wang2020b}, for eight shape categories selected by \cite{Wang2020b}.}
%    \scalebox{0.95}{
%        \scriptsize
        \begin{tabular}{c|cc}
            \toprule
            \thead{Category} & \thead{~\cite{Wang2020b}} & \thead{Ours}  \\
            \midrule
            Airplane         & 67.3                      & \textbf{73.9} \\
            Bag              & 74.4                      & \textbf{81.7} \\
            Cap              & \textbf{86.3}             & 84.4          \\
            Chair            & 83.4                      & \textbf{87.2} \\
            Lamp             & 68.7                      & \textbf{76.5} \\
            Laptop           & 93.8                      & \textbf{95.4} \\
            Mug              & 90.9                      & \textbf{95.5} \\
            Table            & 74.2                      & \textbf{74.8} \\
            \midrule
            Mean             & 79.8                      & \textbf{83.7} \\
            \bottomrule
        \end{tabular}
%    }\vspace{2mm}
    \label{tab:shapenetpart-10-samples}
    \vspace{-3mm}
\end{table}

\subsubsection{ShapeNetPart segmentation}\label{subsubsec:shapenetpart}
As there is no structural hierarchy in ShapeNetPart,  hierarchy consistency loss was dropped from our loss function. We report the mean IoU across all categories (c-mIoU) and across all instances (i-mIoU),  commonly used metrics in the ShapeNetPart segmentation benchmark. \cref{tab:shapenetpart-comp} compares results from various methods using 5\% labeled data for training. It is clear that our method is superior to others, while {MIDNet} is  second best. We also made a more thorough comparison to {MIDNet} using other ratios of labeled training data. 
The results in \cref{tab:shapenetpart-eval} show that {MCNet} always performed much better than {MIDNet}.

We also conducted a few-shot experiment by following the setting in the state-of-the-art few-shot 3D segmentation method \cite{Wang2020b}: eight categories of ShapeNetPart were tested and 10 labeled shapes used for training. We trained our network 5 times and  sampled 10 labeled shapes randomly from the original dataset each time, reporting  average results in
\cref{tab:shapenetpart-10-samples}. It shows that our method is superior to that of \cite{Wang2020b} for most of the tested categories.

%%%%%%%%%%%%%%%%%%%%%%%%%%%%%%%%%%%%%%%%%%%%%%%%%%%%%%%%
\subsection{ScanNet segmentation}\label{subsubsec:scannet}
\subsubsection{Setting}
We have also applied our semi-supervised method to indoor semantic scene segmentation. We choose the ScanNet dataset~\cite{dai2017scannet} as a testbed, and used 1\%, 5\%, 10\%, and 20\% labeled scenes  from the original training dataset, with the remainder of the training dataset  regarded as unlabeled data. For a fair comparison, we used the labeled scenes from \cite{Hou2021}.  We measured mean IoU to evaluate  segmentation quality on the validation set.  As the ScanNet dataset does not provide hierarchical segmentation, we defined a 2-level segmentation on ScanNet: classes at the fine level are the original segmentation classes; at the coarse level, we merged the semantic classes into 6 categories using semantic affinity according to WordNet~\cite{Wordnet}, calling this hierarchy HW. Details of these hierarchies are presented in Appendix B. As our part substitution is not intended for 3D scenes, we did not synthesize 3D labeled scenes for training.

\begin{table}[b]
    \centering
    \caption{Fine level segmentation results for ScanNet. \textbf{r} is the proportion of labeled data used for  training. $-$H means without hierarchy, $+$H means with hierarchy. The  model from \cite{Hou2021} was used to generate its segmentation results.
    }
%    \scalebox{1}{ \scriptsize
        \begin{tabular}{cc|c}
            \toprule
            \thead{r}               & \thead{Method}                                                                  & \thead{mIoU} \\
            \midrule
            \multirow{4}{*}{$1\%$}  & \cite{Hou2021}                                                                  & 29.3                   \\
                                    & Our supervised baseline                                                         & 27.0                   \\
                                    & {MCNet}$-$H  $(\lambda_\mathrm{pts}= \lambda_\mathrm{part}=0.005)$           & 28.7                   \\
                                    & {MCNet}$+$H $(\lambda_\mathrm{pts}= \lambda_\mathrm{part}=\lambda_{h}=0.005)$ & \textbf{29.4}          \\
            \midrule
            \multirow{4}{*}{$5\%$}  & \cite{Hou2021}                                                                  & 45.4                   \\
                                    & Our supervised baseline                                                         & 47.9                   \\
                                    & {MCNet}$-$H  $(\lambda_\mathrm{pts}= \lambda_\mathrm{part}=0.05)$            & 48.2                   \\
                                    & {MCNet}$+$H $(\lambda_\mathrm{pts}= \lambda_\mathrm{part}=\lambda_{h}=0.05)$  & \textbf{48.3}          \\
            \midrule
            \multirow{4}{*}{$10\%$} & \cite{Hou2021}                                                                  & 59.5                   \\
                                    & Our supervised baseline                                                         & 58.1                   \\
                                    & {MCNet}$-$H  $(\lambda_\mathrm{pts}= \lambda_\mathrm{part}=0.1)$             & 59.1                   \\
                                    & {MCNet}$+$H $(\lambda_\mathrm{pts}= \lambda_\mathrm{part}=\lambda_{h}=0.1)$   & \textbf{60.3}          \\
            \midrule
            \multirow{4}{*}{$20\%$} & \cite{Hou2021}                                                                  & 64.1                   \\
                                    & Our supervised baseline                                                         & 62.8                   \\
                                    & {MCNet}$-$H  $(\lambda_\mathrm{pts}= \lambda_\mathrm{part}=0.1)$             & 63.9                   \\
                                    & {MCNet}$+$H $(\lambda_\mathrm{pts}= \lambda_\mathrm{part}=\lambda_{h}=0.1)$   & \textbf{64.9}          \\
            \bottomrule
        \end{tabular}
%    }\vspace{2mm}
    \label{tab:scannet-eval}
    %\vspace{-3mm}
\end{table}

\subsubsection{Data perturbation}
We used the same augmentation configuration as~\cite{Hou2021}: a random rotation with pitch, yaw, and roll angles  less than $\ang{3}, \ang{180}, \ang{3}$, respectively, a uniform scaling in the range $[0.9, 1.1]$,  random translations along $x$-, $y$- axes within the range $[0.8, 1.2]$, and a color transformation including auto contrast, color translation and color jitter. We randomly sampled 20\% points from a scene in each training iteration.

\subsubsection{Parameters and training protocol} Each batch contained 4 shapes, with two from the labeled scenes and two from the unlabeled scenes. A maximum of 60000 iterations was used.  We used the SGD optimizer with a learning rate of 0.1, decayed at the 30000-th  and 45000-th iterations by a factor of 0.1. We tried different multilevel consistency weights, and found that smaller weights improve results when the proportion of labeled data is low. The optimal settings we found are reported alongside the network results in \cref{tab:scannet-eval}.

\subsubsection{Results} We compared our supervised baseline, \ie using the segmentation loss and  labeled data only, our method with and without  structural hierarchy loss, and the state-of-the-art unsupervised pretraining with fine-tuning method proposed by \cite{Hou2021}. As   \cref{tab:scannet-eval} shows, our supervised baseline and our method without using the hierarchy loss worked less well than \cite{Hou2021} except in the test with 5\% labeled data. With the additional hierarchy loss, our method performed best in all tests.

\subsubsection{Sensitivity to customized hierarchy}
To study whether our method on ScanNet is sensitive to the customized hierarchy, we randomly grouped fine level parts into 6 categories three times, and created three different 2-level hierarchies, HA, HB, HC. \cref{tab:scannet-hierarchy} reports the segmentation results using  these customized hierarchies. We find that {MCNet} achieves similar results using HW, HA, HB, and HC, so conclude that while our approach benefits from  hierarchical relationships,  it is insensitive to the hierarchy construction.

\begin{table}[h]
    \centering
    \caption{Fine level segmentation results for ScanNet using a different customized hierarchy. \textbf{r} is the proportion of labeled data used for  training.}
%    \scalebox{0.95}{ \scriptsize
        \begin{tabular}{c|c|c|c|c}
            \toprule
            \thead{r} & \thead{HW} & \thead{HA} & \thead{HB} & \thead{HC} \\
            \midrule
            $1\%$     & 29.4       & 29.2       & 29.7       & 29.5       \\
            % \midrule
            $5\%$     & 48.3       & 48.9       & 48.6       & 48.4       \\
            % \midrule
            $10\%$    & 60.3       & 60.2       & 60.2       & 60.0       \\
            % \midrule
            $20\%$    & 64.9       & 64.8       & 64.4       & 64.7       \\
            \bottomrule
        \end{tabular}
%    }\vspace{2mm}
    \label{tab:scannet-hierarchy}
    %\vspace{-3mm}
\end{table}

%%%%%%%%%%%%%%%%%%%%%%%%%%%%%%%%%%%%%%%%%%%%%%%%%%%%%%%%%%%%%%%%%%%%%%%%%%%%%%%%%%%

%%%%%%%%%%%%%%%%%%%%%%%%%%%%%%%%%%%%%%%%%%%%%%%%%%%%%%%%
\subsection{Ablation study} \label{subsec:ablation}
We next evaluate the efficacy of our consistency loss, part substitution, and hyper-parameter selection approaches, using the four biggest categories from PartNet: Chair, Table, Storage, and Lamp, as our testbed.  We used 2\% labeled data here only. Results are reported in \cref{tab:partnet-data}.  The network was trained for each shape category individually.

\begin{table}[h]
    \centering
    \caption{Numbers of shapes in the four shape categories from PartNet used in our ablation study.  }
%    \scalebox{1}{ \scriptsize
        \begin{tabular}{l*{4}{r}}
            \toprule
                                   & \thead{Chair} & \thead{Lamp} & \thead{Storage} & \thead{Table} \\\midrule
            \bfseries Labeled   & 90            & 31           & 32            & 114            \\
            \bfseries Unlabeled & 4399          & 1523         & 1556          & 5593          \\
            \bfseries Test      & 1217          & 419          & 451           & 1668          \\
            \bottomrule
        \end{tabular}%}
    \label{tab:partnet-data} %\vspace{-2mm}
\end{table}

\subsubsection{Multilevel consistency loss and part substitution} \label{subsubsec:ablationloss}
We designed a series of ablation studies to validate the advantage of our multilevel consistency losses and part substitution, using as baseline  our network trained with the multilevel segmentation loss on the limited labeled data only. \cref{tab:lossablation} reports results for the baseline (ID-(1)) and the baseline with different combinations of our multilevel consistency losses with semi-supervised training. We state how many   labeled shapes were synthesized via multilevel part substitution from the 2\% labeled data  used for training; $N$ indicates the total number of labeled and unlabeled data items.

Experiments (2)--(4) clearly show that utilization of any consistency loss can improve segmentation accuracy. Experiment (5) indicates that synthesizing labeled shapes by part substitution can significantly improve the network results even when used in a purely supervised training manner. Combinations of different types of consistency losses (6)--(9) further boost network performance; combining all consistency losses in (9) works best. Adding synthesized labeled shapes (10)--(12) further helps the network to reach its highest accuracy. The configuration in (12) is the default and optimal setting of {MCNet} used in \cref{subsec:shapeseg}.

Concurrent work to this work, PointCutMix~\cite{zhang2021pointcutmix} proposes a data augmentation method which finds the optimal assignment between two labeled point clouds and generates new training data by replacing points in one sample with their optimally assigned pairs. We implemented their approach and used the generated shapes to enhance  training. A few synthesized shapes are illustrated in Appendix C. Experiments (13) and (14) show that their data augmentation method can enhance the network accuracy to a certain degree, but does not bring as significant an improvement as our approach, due to its lack of awareness of part structures during data synthesis.

\begin{table*}[t]
    \centering
    \caption{
        Ablation study for {MCNet} trained on four categories from PartNet using different loss combinations and synthesized shapes, and 2\% labeled data. \checkmark\; indicates that the corresponding loss was employed during  training.  Aug is the number of synthesized shapes used. $N$ is  the total number of labeled and unlabeled data items. In (13), (14), we used the method of \cite{zhang2021pointcutmix} to generate augmented labeled shapes for training.  Quality metrics were measured on the test dataset.%
        \label{tab:lossablation}
    }
%    \scalebox{0.95}{ \scriptsize
\setlength{\tabcolsep}{1mm}
        \begin{tabular}{cccccr|cc|cc|cc|cc}
            \toprule
            \multicolumn{6}{c}{\bfseries Experimental configuration} & \multicolumn{2}{c}{\bfseries Coarse Level} & \multicolumn{2}{c}{\bfseries Medium Level} & \multicolumn{2}{c}{\bfseries Fine Level} & \multicolumn{2}{c}{\bfseries Avg}                                                                                                                                                                           \\
            \midrule

            \thead{ID}                                             & \thead{$L_\texttt{seg}$}                   & \thead{$L_\texttt{point}$}                 & \thead{$L_\texttt{part}$}                & \thead{$L_\texttt{h}$}            & \thead{Aug}                 & \thead{p-mIoU} & \thead{s-mIoU} & \thead{p-mIoU} & \thead{s-mIoU} & \thead{p-mIoU} & \thead{s-mIoU} & \thead{p-mIoU} & \thead{s-mIoU} \\
            \midrule
            (1)                                                    & \checkmark                                 &                                            &                                          &                                   & 0                               & 37.7           & 56.2           & 25.6           & 34.0           & 21.8           & 30.3           & 28.3           & 40.2           \\
            (2)                                                    & \checkmark                                 & \checkmark                                 &                                          &                                   & 0                               & 40.3           & 59.8           & 27.6           & 37.1           & 23.4           & 33.6           & 30.4           & 43.5           \\
            (3)                                                    & \checkmark                                 &                                            & \checkmark                               &                                   & 0                               & 40.5           & 57.9           & 26.4           & 35.3           & 22.5           & 31.1           & 29.8           & 41.4           \\
            (4)                                                    & \checkmark                                 &                                            &                                          & \checkmark                        & 0                               & 39.9           & 59.1           & 27.1           & 37.7           & 23.3           & 33.6           & 30.1           & 43.5           \\
            (5)                                                    & \checkmark                                 &                                            &                                          &                                   & $N$                             & 41.2           & 62.1           & 28.7           & 40.4           & 24.2           & 36.2           & 31.3           & 46.2           \\
            (6)                                                    & \checkmark                                 & \checkmark                                 & \checkmark                               &                                   & 0                               & 41.3           & 61.3           & 27.7           & 39.8           & 23.6           & 35.9           & 30.9           & 45.6           \\
            (7)                                                    & \checkmark                                 & \checkmark                                 &                                          & \checkmark                        & 0                               & 40.5           & 62.4           & 27.7           & 40.9           & 23.5           & 36.9           & 30.6           & 46.7           \\
            (8)                                                    & \checkmark                                 &                                            & \checkmark                               & \checkmark                        & 0                               & 41.3           & 60.6           & 27.6           & 38.6           & 23.5           & 34.6           & 30.8           & 44.6           \\
            (9)                                                    & \checkmark                                 & \checkmark                                 & \checkmark                               & \checkmark                        & 0                               & 42.7           & 62.7           & 28.4           & 41.5           & 24.2           & 37.5           & 31.7           & 47.2           \\
            (10)                                                   & \checkmark                                 & \checkmark                                 & \checkmark                               & \checkmark                        & $N/4$                           & 42.8           & 64.7           & 29.7           & 43.7           & 26.0           & 39.5           & 32.8           & 49.3           \\
            (11)                                                   & \checkmark                                 & \checkmark                                 & \checkmark                               & \checkmark                        & $N/2$                           & 42.8           & 65.3           & 30.3           & 44.1           & 26.1           & \textbf{39.9}  & 33.1           & 49.8           \\
            (12)                                                   & \checkmark                                 & \checkmark                                 & \checkmark                               & \checkmark                        & $N$                             & \textbf{43.1}  & \textbf{65.6}  & \textbf{30.5}  & \textbf{44.2}  & \textbf{26.3}  & \textbf{39.9}  & \textbf{33.3}  & \textbf{49.9}  \\
            \midrule
            (13)                                                   & \checkmark                                 &                                            &                                          &                                   & $N$ \cite{zhang2021pointcutmix} & 38.2           & 54.6           & 27.7           & 35.0           & 23.4           & 31.6           & 29.8           & 40.4           \\
            (14)                                                   & \checkmark                                 & \checkmark                                 & \checkmark                               & \checkmark                        & $N$ \cite{zhang2021pointcutmix} & 40.4           & 59.8           & 30.1           & 39.9           & 24.9           & 36.2           & 31.8           & 45.3           \\
            \bottomrule
        \end{tabular}
%    }
    %\vspace{-2mm}
\end{table*}

\subsubsection{Data perturbation} \label{subsubsec:perturbation}

We also examined how data perturbation affects the performance of {MCNet}.  The experimental setup was as in \cref{subsubsec:ablationloss}; we only varied the ranges of  rotation, scaling, and translation data perturbation parameters, with results shown in \cref{tab:perturbation-ablation}.
Configuration (1) is our default configuration. 

By varying the range of the random rotation angle, we found that when using a smaller or larger angle range, the network performance slightly decreases: see (1)--(3).

Tests (1), (4), and (5) reveal that an appropriate shape scaling is important. Note that any part outside the unit sphere caused by a large scaling is removed by our perturbation, so there are fewer corresponding points between the two perturbed shape copies used in our consistency loss computation.

Tests (1), (6), and (7) also show that an appropriate translation helps our training. Making translation too large also results in missing  shape geometry, degrading the efficacy of our consistency loss.

\begin{table*}[t]
    \centering
    \caption{
        Ablation study for {MCNet} trained on four categories from PartNet under different data perturbation configurations, with 2\% labeled data. Quality metrics were measured on the test dataset. \label{tab:perturbation-ablation}
    }
%    \scalebox{0.95}{\scriptsize
        \setlength{\tabcolsep}{1mm}
        \begin{tabular}{cccc|cc|cc|cc|cc}
            \toprule
            \multicolumn{4}{c}{\bfseries Perturbation configuration} & \multicolumn{2}{c}{\bfseries Coarse Level} & \multicolumn{2}{c}{\bfseries Medium Level} & \multicolumn{2}{c}{\bfseries Fine Level} & \multicolumn{2}{c}{\bfseries Avg}                                                                                                                        \\
            \midrule
            \thead{ID}                                               & \thead{rotation}                           & \thead{scaling}                            & \thead{translation}                      & \thead{p-mIoU}                    & \thead{s-mIoU} & \thead{p-mIoU} & \thead{s-mIoU} & \thead{p-mIoU} & \thead{s-mIoU} & \thead{p-mIoU} & \thead{s-mIoU} \\
            \midrule
            (1)                                                      & [-\ang{10}, \ang{10}]                      & [0.75, 1.25]                               & [-0.25, 0.25]                            & 43.1                              & \textbf{65.6}  & \textbf{30.5}  & \textbf{44.2}  & \textbf{26.3}  & \textbf{39.9}  & \textbf{33.3}  & \textbf{49.9}  \\
            (2)                                                      & [\;-\ang{5},\;\;\, \ang{5}]                & [0.75, 1.25]                               & [-0.25, 0.25]                            & 43.0                              & 65.4           & 30.4           & 43.8           & 25.9           & 39.6           & 33.1           & 49.6           \\
            (3)                                                      & [-\ang{20}, \ang{20}]                      & [0.75, 1.25]                               & [-0.25, 0.25]                            & 41.6                              & \textbf{65.6}  & \textbf{30.5}  & 43.6           & 26.1           & 39.8           & 32.7           & 49.7           \\
            \midrule
            (4)                                                      & [-\ang{10}, \ang{10}]                      & [0.90, 1.10]                               & [-0.25, 0.25]                            & 42.7                              & 64.5           & 29.9           & 43.1           & 25.7           & 39.1           & 32.7           & 48.9           \\
            (5)                                                      & [-\ang{10}, \ang{10}]                      & [0.60, 1.40]                               & [-0.25, 0.25]                            & \textbf{43.3}                     & 64.8           & 30.3           & 43.4           & 26.0           & 39.5           & 33.2           & 49.2           \\
            \midrule
            (6)                                                      & [-\ang{10}, \ang{10}]                      & [0.75, 1.25]                               & [-0.10, 0.10]                            & 41.5                              & 64.8           & 29.7           & 43.4           & 25.5           & 39.3           & 32.2           & 49.2           \\
            (7)                                                      & [-\ang{10}, \ang{10}]                      & [0.75, 1.25]                               & [-0.40, 0.40]                            & 41.7                              & 64.6           & 29.8           & 43.1           & 25.6           & 39.2           & 32.3           & 49.0           \\

            \bottomrule
        \end{tabular}
%    }
    %\vspace{-2mm}
\end{table*}

\section{Conclusions} \label{sec:conclusion}
We have presented an effective semi-supervised approach for 3D shape segmentation. Our novel multilevel consistency and part substitution scheme harnesses the structural consistency hidden in both unlabeled data and labeled data, for our network training,  leading to  superior performance on 3D segmentation tasks with few labeled data items. We believe that our multilevel consistency will find more applications, potentially being useful for semi-supervised image segmentation.

There are still a few unexplored directions. Firstly, it is possible to extend the hierarchical consistency from points to parts and involve more structural levels ($> 3$) to improve the training, which may be especially beneficial for more complicated datasets and part structures. Secondly, it would be helpful to synthesize novel shapes and scenes from both labeled and unlabeled data with more diverse structural and geometry variations for semi-supervised learning.

\bibliographystyle{unsrt}  
\bibliography{references}  

\clearpage

\section*{Appendices}
\subsection*{A. Network structure for scene segmentation}
\begin{figure}[h]
    \centering
    \includegraphics[width=0.7\columnwidth]{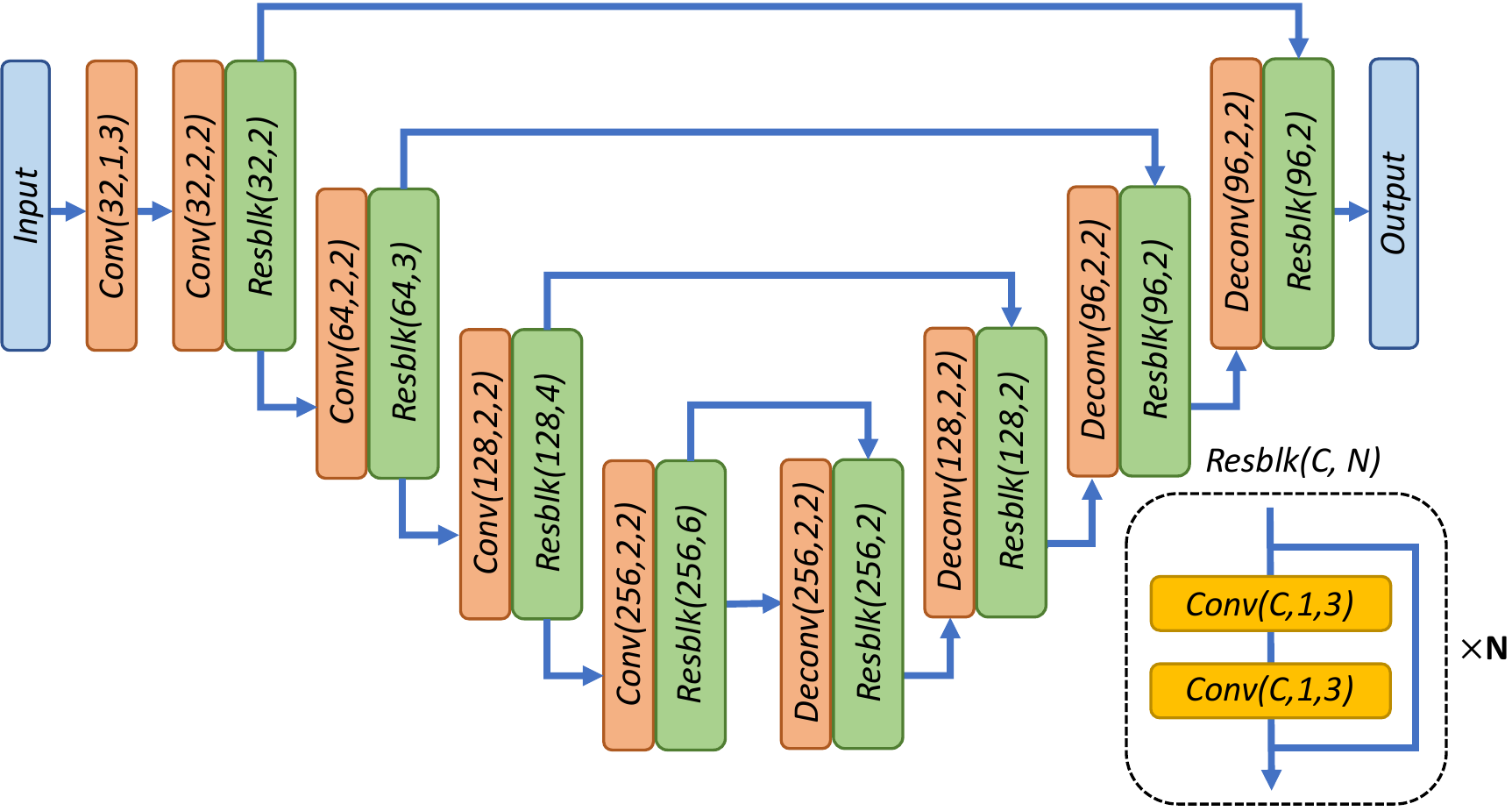}
    \caption{Octree-based U-Net structure for scene segmentation. Conv(C, S, K) and Deconv(C, S, K) represent octree-based convolution and deconvolution. C, S, K are the number of output channels, stride, and kernel size.}
    \label{fig:unet_scene} %\vspace{-2mm}
\end{figure}

An octree-based U-Net structure is used as our base network. It has four levels of domain resolution: see \cref{fig:unet_scene}. The maximum octree depth is \num{9}.

\subsection*{B. ScanNet hierarchy}
The coarse levels of HW, HA, HB, and HC are shown in  \cref{tab:scannet-hierarchy-definition}. Fine classes are merged to the coarse level.  Numbers in the table give the coarse label ID.

\begin{table}[h]
    \centering
    \caption{Customized hierarchies for ScanNet. The number is the coarse label ID.}
%    \scalebox{1}{ \scriptsize
        \begin{tabular}{c|c|c|c|c}
            \toprule
            \thead{Category} & \thead{HW} & \thead{HA} & \thead{HB} & \thead{HC} \\
            \midrule
            Wall             & 1          & 6          & 4          & 4          \\
            Floor            & 2          & 3          & 6          & 3          \\
            Cabinet          & 3          & 5          & 4          & 2          \\
            Bed              & 3          & 3          & 2          & 5          \\
            Chair            & 3          & 2          & 3          & 3          \\
            Sofa             & 3          & 4          & 5          & 4          \\
            Table            & 3          & 3          & 3          & 5          \\
            Door             & 4          & 4          & 5          & 1          \\
            Window           & 1          & 2          & 6          & 4          \\
            Bookshelf        & 4          & 2          & 5          & 5          \\
            Picture          & 4          & 1          & 6          & 2          \\
            Counter          & 3          & 2          & 6          & 1          \\
            Desk             & 3          & 2          & 5          & 1          \\
            Curtain          & 5          & 1          & 2          & 2          \\
            Refrigerator     & 4          & 1          & 4          & 2          \\
            Shower curtain   & 5          & 2          & 1          & 2          \\
            toilet           & 6          & 4          & 4          & 1          \\
            sink             & 6          & 6          & 6          & 2          \\
            bathtub          & 5          & 5          & 6          & 5          \\
            Other furniture  & 4          & 1          & 3          & 2          \\

            \bottomrule
        \end{tabular}
%    }\vspace{2mm}
    \label{tab:scannet-hierarchy-definition}
%    \vspace{-3mm}
\end{table}

\subsection*{C. Data augmentation by part substitution}
\begin{figure*}[h]
    \centering
    \begin{overpic}[width=1\linewidth]{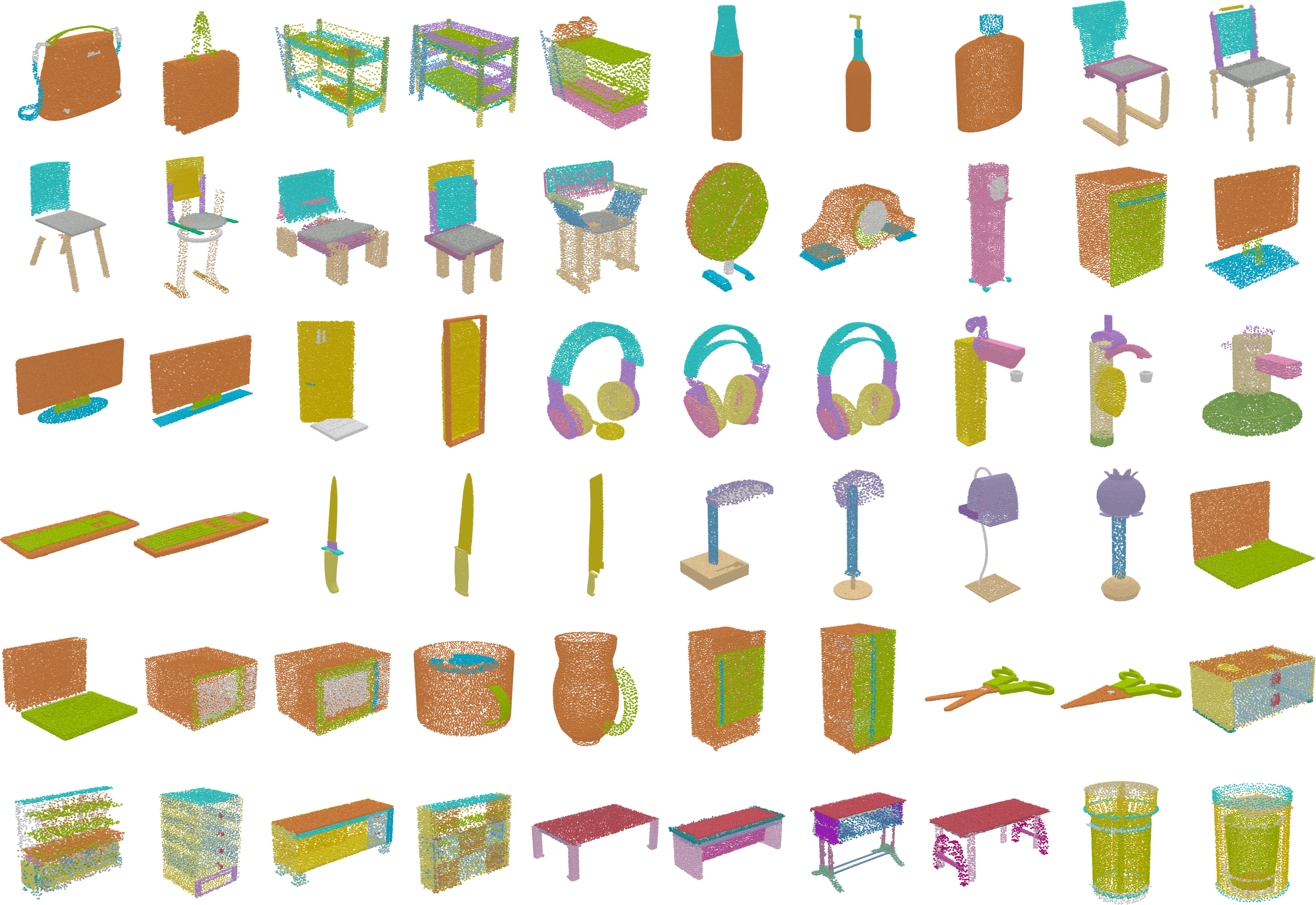}
    \end{overpic}
    \caption{Augmented shapes based on 2\% labeled data from PartNet.}
    \label{fig:augmentation_partnet}
\end{figure*}

\begin{figure*}[h]
    \centering
    \begin{overpic}[width=1\linewidth]{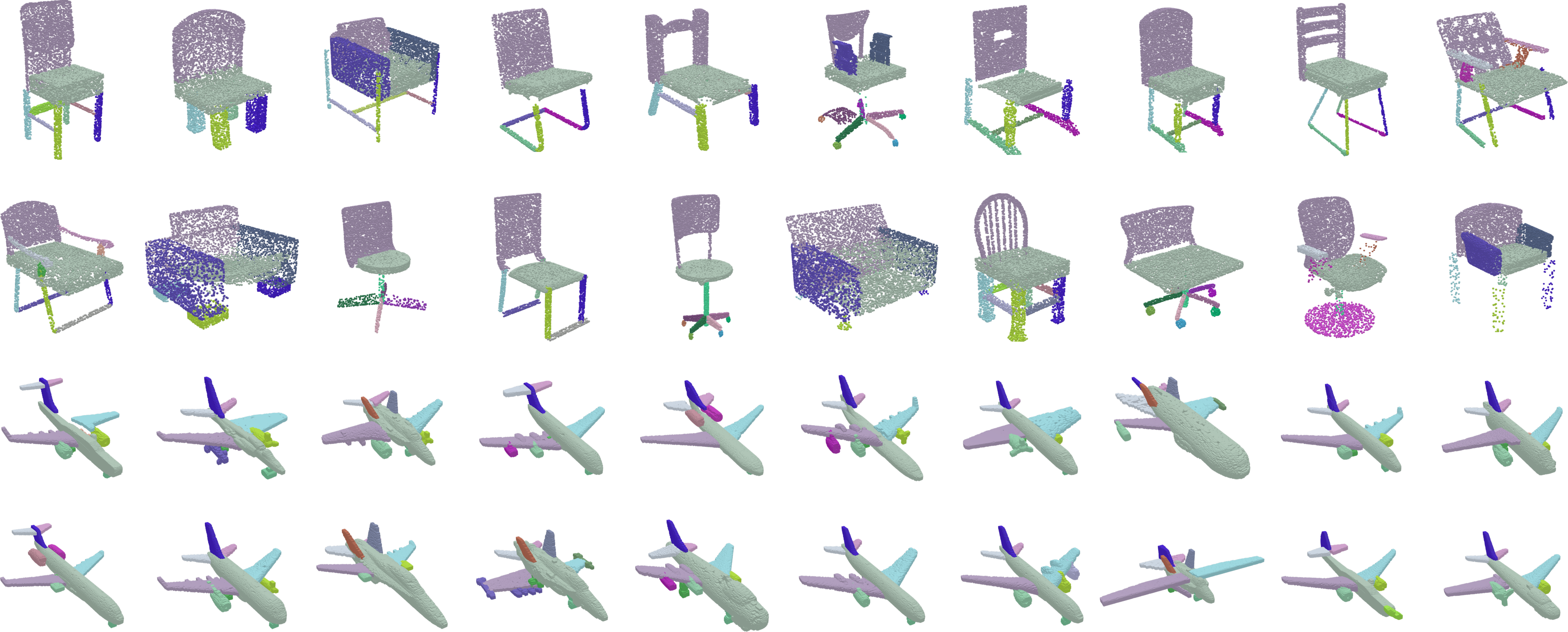}
    \end{overpic}
    \caption{Augmented shapes based on 2\% labeled data from Chair2 and Airplane.}
    \label{fig:augmentation_ourdata}
\end{figure*}

\begin{figure*}[t]
    \centering
    \begin{overpic}[width=1\linewidth]{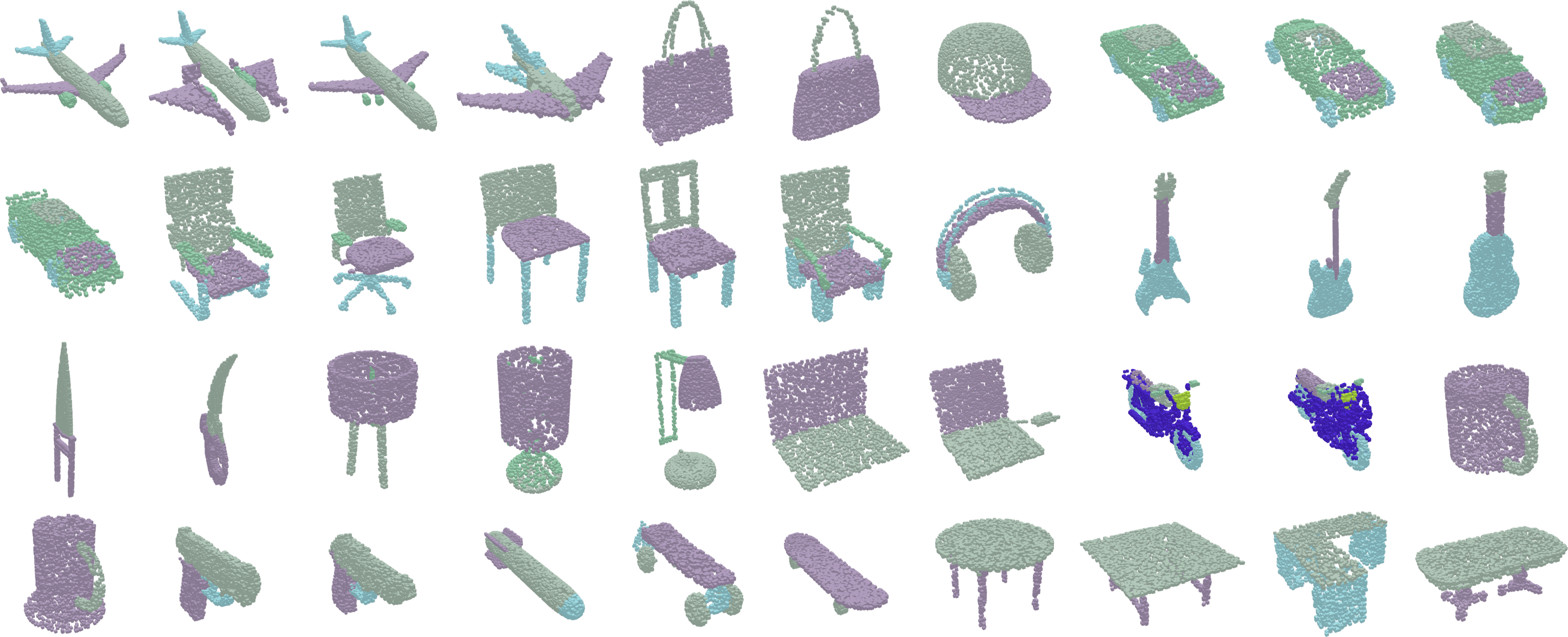}
    \end{overpic}
    \caption{Illustration of the augmented shapes based on 2\% labeled data of ShapeNetPart.}
    \label{fig:augmentation_shapenetpart}
\end{figure*}

\begin{figure*}[t]
    \centering
    \begin{overpic}[width=1\linewidth]{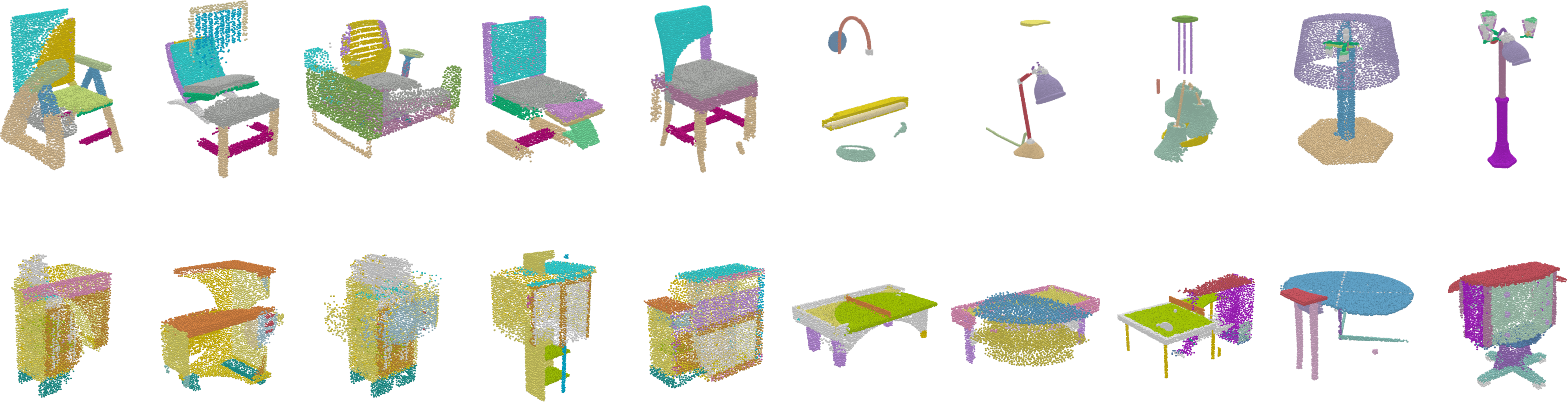}
    \end{overpic}
    \caption{Augmented shapes based on 2\% labeled data from PartNet using PointCutMix~\cite{zhang2021pointcutmix}.}
    \label{fig:augmentation_pointcutmix}
\end{figure*}

In \cref{fig:augmentation_ourdata,fig:augmentation_partnet,fig:augmentation_shapenetpart}, we illustrate a sample set of shapes augmented by part substitution on 2\% labeled data. The majority of the augmented shapes are plausible and would help to enrich the labeled data for network training. In \cref{fig:augmentation_pointcutmix}, we render the augmented shapes generated by the approach of \cite{zhang2021pointcutmix}. We can observe many implausible shape parts which do not assist training.

\end{document}